\documentclass{article}

\usepackage{microtype}
\usepackage{graphicx}
\usepackage{subfigure}
\usepackage{booktabs} % for professional tables

\usepackage{hyperref}

\usepackage[accepted]{icml2021}

\usepackage[utf8]{inputenc} % allow utf-8 input
\usepackage[T1]{fontenc}    % use 8-bit T1 fonts
\usepackage{hyperref}       % hyperlinks
\usepackage{url}            % simple URL typesetting
\usepackage{amsmath}
\usepackage{amsfonts}       % blackboard math symbols
\usepackage{nicefrac}       % compact symbols for 1/2, etc.

\usepackage{comment}
\usepackage{pdflscape}
\usepackage{algorithm}
\usepackage{algorithmic}
\usepackage{bm}
\usepackage{soul}
\usepackage{multirow}

\usepackage{float}

\icmltitlerunning{Gaussian Process Inference Networks for VAE}

\begin{document}

\onecolumn
%\twocolumn[
\icmltitle{Reducing the Amortization Gap in Variational Autoencoders:\\
A Bayesian Random Function Approach}




%\vspace{-1.0em}

\begin{icmlauthorlist}
\icmlauthor{Minyoung Kim}{sa}
\icmlauthor{Vladimir Pavlovic}{sa,ru}
\end{icmlauthorlist}

\icmlaffiliation{sa}{Samsung AI Center  Cambridge, UK}
\icmlaffiliation{ru}{Rutgers University,
  Piscataway, NJ, USA}
\icmlcorrespondingauthor{Minyoung Kim}{mikim21@gmail.com}

\icmlkeywords{Machine Learning, ICML}

\vskip 0.3in
%]



\printAffiliationsAndNotice{}  % leave blank if no need to mention equal contribution

%\vspace{-1.0em}

\begin{abstract}
Variational autoencoder (VAE) is a very successful generative model %in which a highly nonlinear deep generative process can be easily incorporated. Especially, the VAE enjoys 
%that benefits from computational advantage from 
whose key element is the so called amortized inference network, which can perform test time inference using a single feed forward pass. %This allows one to circumvent time-consuming variational optimization for individual data instances at test time. 
Unfortunately, this comes at the cost of degraded accuracy in posterior approximation, often underperforming the instance-wise variational optimization. Although the latest semi-amortized approaches mitigate the issue by performing a few variational optimization updates starting from the VAE's amortized inference output, they inherently suffer from computational overhead for inference at test time. In this paper, we address the problem in a completely different way by considering a random inference model, where we model the mean and variance functions of the variational posterior as random Gaussian processes (GP). %, and perform the posterior inference within the Bayesian GP framework. 
The motivation is that the deviation of the VAE's amortized posterior distribution from the true posterior can be regarded as random noise, which allows us to take into account the uncertainty in posterior approximation in a principled manner. In particular, our model can quantify the difficulty in posterior approximation by a Gaussian variational density. %that might be different from instance to instance.
Inference in our GP model is done by a single feed forward pass through the network, significantly faster than semi-amortized methods. We show that our approach attains higher test data likelihood than the state-of-the-arts %and even the high-capacity flow-based encoder models
on several benchmark datasets. 
\end{abstract}

%%%%%%%%%%%%%%%%%%%%%%%%%%%%%%%%%%%%%%%%%%%%%%%%%%%%%%%%%%%%%%%%%%%%%%%%%%%%%%%
%%%%%%%%%%%%%%%%%%%%%%%%%%%%%%%%%%%%%%%%%%%%%%%%%%%%%%%%%%%%%%%%%%%%%%%%%%%%%%%
\section{Introduction}\label{sec:intro}

%\hl{Make it brief and informal, like 4 paragraphs in 1 page; Instead, add Background section for full formal discussion}

%1) Intro to VAE: Why it is successful and the benefit (computationally, due to amortized inference network)
Variational Autoencoder (VAE)~\cite{vae14,vae14r} is a very successful generative model where a highly complex deep nonlinear generative process can be easily incorporated. A key element of the VAE, the deep inference (a.k.a. encoder) network, can perform the test time inference using a single feed forward pass through the network, bringing significant computational speed-up. This feature, known as {\em amortized inference}, allows the VAE to circumvent otherwise time-consuming steps of solving the variational optimization problem for each individual instance at test time, required in the standard variational inference techniques, such as the stochastic variational inference (SVI)~\cite{svi}. 
%Compared to the standard variational inference techniques, such as the stochastic variational inference (SVI)~\cite{svi}, in which one has to solve the variational optimization problem for each individual instance at test time, VAE avoids this time consuming task. 

%2) Recent study on amortization error (Cremer) -- inaccurate compared to the SVI. But SVI is slow. Perhaps soemthing in between is the semi-amortized approaches proposed but incur additional issues (computational overhead, difficulty of hyperparam selection)

As suggested by the recent study~\cite{cremer18}, however, the amortized inference can also be a drawback of the VAE, specifically the accuracy of posterior approximation by the amortized inference network is often lower than the accuracy of the SVI's full variational optimization. There are two general approaches to reduce this amortization error. The first is to increase the network capacity of the inference model (e.g., flow-based models~\cite{iafvae,hfvae}). The other direction is the so-called {\em semi-amortized} approach~\cite{savae_ykim,savae_krishnan,savae_marino,vlae}, where the key idea is to use the VAE's amortized inference network to produce a good initial distribution, from which a few SVI steps are performed at test time to further reduce the amortization error, quite similar in nature to the test time model adaptation of the MAML~\cite{maml} in multi-task (meta) learning. Although these models often lead to improved posterior approximation, they raise several issues: Training the models for the former family of approaches is usually difficult because of the increased model complexity;  the latter approaches inadvertently suffer from computational overhead of additional SVI gradient steps at test time. 

%3) Our approach (we stick to amortized inference model for its computational merit, but regard distribution's means and variances as random functions) and motivation. The first motivation comes from the vae's suboptimality, namely single point estimate is not accurate, thus we have to incorporate uncertainty in Bayesian framework. Second motivation ...
In this paper, we propose a novel approach to address these drawbacks. We retain the amortized inference framework similar to the standard VAE for its computational benefits, but consider a {\em random inference model}. Specifically, the mean and the variance functions of the variational posterior distribution are  a priori assumed to be Gaussian process (GP) distributed. There are two main motivations for this idea. 
The first one stems from the suboptimality of the VAE, where the estimated amortized inference network suffers from deviation from the true posteriors. This inaccuracy can be characterized by inherent uncertainty in the posterior approximation of the deterministic amortized inference network, suggesting the need for a principled Bayesian uncertainty treatment. 
The second intuition is that the deviation of the VAE's variational posterior distributions from the true posteriors can be naturally regarded as random noise. Whereas the semi-amortized approaches perform extra SVI gradient updates at test time to account for this noise, we model the discrepancy using a Bayesian neural network (GP), resulting in a faster and more accurate amortized model via principled uncertainty marginalization. Another benefit of the Bayesian treatment is that our model can quantify the discrepancy in approximation, which can serve as useful indicators for goodness of posterior approximations.

The inference in our model is significantly faster than that of semi-amortized methods, accomplished by a single feed forward pass through the GP posterior marginalized inference network. We show that our approach attains higher test data likelihood scores than the state-of-the-art semi-amortized approaches and even the high-capacity flow-based encoder models on several benchmark datasets.

%%%%%%%%%%%%%%%%%%%%%%%%%%%%%%%%%%%%%%%%%%%%%%%%%%%%%%%%%%%%%%%%%%%%%%%%%%%%%%%
%%%%%%%%%%%%%%%%%%%%%%%%%%%%%%%%%%%%%%%%%%%%%%%%%%%%%%%%%%%%%%%%%%%%%%%%%%%%%%%
\section{Background}\label{sec:background}

%\hl{Just excerpted from recursive mixture, needs to be rephrased.}

%%%%%%%%%%%%%%%%%%%%%%%%%%%%%%%%%%%%%%%%%%%%%%%%%%%%%%%%%%%%%%%%%%%%%%%%%%%%%%%
%\subsection{VAE and Amortized Inference}\label{sec:vae}

Let ${\bf x}\in\mathcal{X}$ be an input data point and ${\bf z}\in\mathbb{R}^d$ be the $d$-variate latent vector. %The %generative process of the 
%VAE is typically 
We consider the generative model\footnote{Although there can be possible variations (e.g., %non-Gaussian prior/conditional, or 
heteroscedastic variance for $p_{\bm{\theta}}({\bf x}|{\bf z})$), %the above is very standard, 
we assume a homoscedastic model for simplicity, and our approach is easily extendable to the variants.}: 
%%%%
\begin{align}
% p({\bf z}) &= \mathcal{N}({\bf z}; {\bf 0}, {\bf I}), \\
% p_{\bm{\theta}}({\bf x}|{\bf z}) &= \mathcal{N}({\bf x}; {\bf g}_{\bm{\theta}}({\bf z}), \sigma_{\bf x}^2{\bf I}),
p({\bf z}) = \mathcal{N}({\bf z}; {\bf 0}, {\bf I}), \ \ \ 
p_{\bm{\theta}}({\bf x}|{\bf z}) = \mathcal{N}({\bf x}; {\bf g}_{\bm{\theta}}({\bf z}), \sigma_{\bf x}^2{\bf I}),
\end{align}
%%%%
where ${\bf g}_{\bm{\theta}}: \mathbb{R}^d \rightarrow \mathcal{X}$ is a (deep) neural network with the weight parameters denoted by $\bm{\theta}$, and $\sigma_{\bf x}^2$ is the variance\footnote{The variance $\sigma_{\bf x}^2$ can be a part of the model to be trained and subsumed in $\bm{\theta}$, but for simplicity we regard it as a fixed constant.} of the white noise. 
For the given data $\mathcal{D} = \{{\bf x}^i\}_{i=1}^N$,  %i.i.d.~samples from the unknown data distribution, 
we maximize the data log-likelihood, $\sum_{i=1}^N \log p_{\bm{\theta}}({\bf x}^i)$, with respect to $\bm{\theta}$ where $p_{\bm{\theta}}({\bf x}) = \mathbb{E}_{p({\bf z)}}[p_{\bm{\theta}}({\bf x}|{\bf z})]$. 
Due to the infeasibility of evaluating the marginal log-likelihood exactly, the variational inference exploits the following inequality,
%%%%
\begin{equation}
\log p_{\bm{\theta}}({\bf x}) \geq 
\mathbb{E}_{q({\bf z}|{\bf x})} \big[ \log p_{\bm{\theta}}({\bf x},{\bf z}) - \log q({\bf z}|{\bf x}) \big], %=: \mathcal{L}(\bm{\theta},\bm{\lambda},{\bf x}),
\label{eq:vae_logpx_elbo}
\end{equation}
%%%%
which holds for {\em any} density $q({\bf z}|{\bf x})$. The inequality becomes tighter as $q({\bf z}|{\bf x})$ becomes closer to the true posterior, as the gap equals $\textrm{KL}(q({\bf z}|{\bf x}) || p_{\bm{\theta}}({\bf z}|{\bf x}))$. 
%\footnote{The gap equals %the KL divergence between $q({\bf z}|{\bf x})$ and the true posterior $p_{\bm{\theta}}({\bf z}|{\bf x})$.
%$\textrm{KL}(q({\bf z}|{\bf x}) || p_{\bm{\theta}}({\bf z}|{\bf x}))$.
%}. 
Then we adopt a tractable density family (e.g., Gaussian) $q_{\bm{\lambda}}({\bf z}|{\bf x})$ parametrized by $\bm{\lambda}$, and maximize the lower bound % (a.k.a.~ELBO) 
in (\ref{eq:vae_logpx_elbo}) w.r.t.~$\bm{\lambda}$. Since our goal is maximizing the log-marginal, $\log p_{\bm{\theta}}({\bf x})$, we also need to optimize the lower bound w.r.t.~$\bm{\theta}$ together with $\bm{\lambda}$, either concurrently or in an alternating fashion. 

Note that at current $\bm{\theta}$, the lower bound optimization w.r.t.~$\bm{\lambda}$ needs to be specific to each input ${\bf x}$, %in principle, 
and hence the optimal solution is dependent on the input ${\bf x}$. Formally, we can denote the optimum by $\bm{\lambda}^*({\bf x})$.
%For instance, for another input point ${\bf x}'$ one should solve the optimization again to find the optimal parameter $\bm{\lambda}'^*$ that best approximates the posterior $p_{\bm{\theta}}({\bf z}|{\bf x}')$. 
%maximizing the lower bound of (\ref{eq:vae_logpx_elbo}) %ELBO w.r.t.~$\bm{\lambda}$ 
The {\em stochastic variational inference} (SVI)~\citep{svi} faithfully implements this idea, and the approximate posterior inference for a new input point ${\bf x}$ in SVI amounts to solving the ELBO optimization on the fly by gradient ascent.
Although this can yield very accurate posterior approximation, it incurs computational overhead since we have to perform full variational optimization for each and every input ${\bf x}$. 
The VAE~\citep{vae14} addresses this problem by introducing a deep neural network $\bm{\lambda}({\bf x}; \bm{\phi})$ with the weight parameters $\bm{\phi}$ as a universal function approximator of the optimum $\bm{\lambda}^*({\bf x})$, and optimize the lower bound w.r.t.~$\bm{\phi}$. This approach is called the {\em amortized variational inference} (AVI). Thus the main benefit of the VAE is computational speed-up as one can simply do feed forward pass through the inference network $\bm{\lambda}({\bf x}; \bm{\phi})$ to perform posterior inference for each ${\bf x}$.

%%%%%%%%%%%%%%%%%%%%%%%%%%%%%%%%%%%%%%%%%%%%%%%%%%%%%%%%%%%%%%%%%%%%%%%%%%%%%%%
%\subsection{Semi-Amortized VAE}\label{sec:savae}

The recent study in~\citep{cremer18} raised the issue of the amortized inference in the VAE, where the quality of data fitting is degraded due to the approximation error between $\bm{\lambda}^*({\bf x})$ and $\bm{\lambda}({\bf x}; \bm{\phi})$, dubbed the {\em amortization error}. To retain the AVI's computational advantage and reduce the amortization error, there were attempts to take the benefits of SVI and AVI, which are referred to as {\em semi-amortized variational inference} (SAVI)~\cite{savae_ykim,savae_marino,savae_krishnan}. The key idea is to learn the amortized inference network to produce a reasonably good initial iterate for the follow-up SVI optimization, perhaps just a few steps. This warm-start SVI gradient ascent would be faster than full SVI optimization, and could reduce the approximation error of the AVI. %'s output\footnote{This is quite similar in nature to the gradient-based meta learning~\cite{maml} that is aimed for fast adaptation of the model to a new task in the multi-task meta learning.}. 

%However, the iterative gradient update procedure in SAVI might be computationally expensive during training since one needs to perform backpropagation for the objective that involves gradients, meaning that Hessian evaluation\footnote{This can be done (memory-)efficiently using the Hessian-vector products with finite difference methods~\citep{hessian_lecun,hessian_domke}, also adopted in~\citep{savae_ykim}.} is needed in essence. 
Although the inference in the SAVI is faster than SVI, it still requires gradient ascent optimization at test time, which might be the main drawback. 
\begin{comment}
\footnote{Also, during training, similar to~\citep{maml}, the objective involves gradients, %update formulas, 
and the backprop essentially requires 
%requiring 
Hessian-vector computation~\citep{hessian_lecun,hessian_domke}. 
%Hessian evaluation. This incurs another computational overhead although there exist approximate finite difference methods for the Hessian-vector products~\citep{hessian_lecun,hessian_domke}.
}. 
\end{comment}
The SAVI also suffers from other minor issues including how to choose the gradient step size and the number of gradient updates to achieve optimal performance-efficiency trade-off\footnote{Although~\citep{vlae} mitigated the issues %of choosing the step size 
by decoder linearization, %with the Laplace approximation, 
%such linearization of the %deep
%decoder network 
it %restricts its applicability to %the models containing 
is rather restricted to 
only fully connected layers, and %makes it 
difficult to be applied to %more structured models such as 
%models like 
convolutional or recurrent networks.}. 
In the next section we propose a novel approach that is much faster than the SAVI, avoiding gradient updates at test time and requiring only feed forward pass through a single network, and at the same time can yield more accurate posterior approximation. %The key idea is to adopt a random Gaussian process (GP) inference model to capture the inherent uncertainty in the approximate posterior estimation. 

%%%%%%%%%%%%%%%%%%%%%%%%%%%%%%%%%%%%%%%%%%%%%%%%%%%%%%%%%%%%%%%%%%%%%%%%%%%%%%%
%%%%%%%%%%%%%%%%%%%%%%%%%%%%%%%%%%%%%%%%%%%%%%%%%%%%%%%%%%%%%%%%%%%%%%%%%%%%%%%
\section{Gaussian Process Inference Network}\label{sec:main}

We start from the variational density of the VAE, but with slightly different notation, as follows:
%%%%
\begin{equation}
q({\bf z}|{\bf x},{\bf f},{\bf h}) = \mathcal{N}\big({\bf z}; {\bf f}({\bf x}), \textrm{Diag}({\bf h}({\bf x}))^2 \big), 
\label{eq:q_z_xfh}
\end{equation}
%%%%
where ${\bf f},{\bf h}:\mathcal{X}\rightarrow \mathbb{R}^d$ are the mean and standard deviation %\footnote{Although we used the term {\em standard deviation}, ${\bf h}({\bf x})$ is not restricted to be a (elementwise) positive function. One can take the absolute %(i.e., the magnitude of ${\bf h}({\bf x})$) to obtain the standard deviation.} 
functions of the variational posterior distribution. %, and $\textrm{Diag}(\cdot)$ forms a diagonal matrix from its input vector. 
Note that if we model ${\bf f}$ and ${\bf h}$ as {\em deterministic} functions (neural networks) and optimize their weight parameters (i.e., point estimation), then it reduces to the standard VAE for which ${\bf f}$ and ${\bf h}$ constitute $\bm{\lambda}({\bf x}; \bm{\phi})$. However, as discussed in the previous sections, such point estimates may be inaccurate, %considerably deviating from the true posterior, 
which implies that there must be inherent uncertainty in posterior approximation. %using a single deterministic amortized inference network. 
To account for the uncertainty, we follow the Bayesian treatment; specifically we let ${\bf f}$ and ${\bf h}$ be independent random GP distributed functions a priori~\cite{gpml_book}, %as follows:
%%%%
\begin{align}
{\bf f}(\cdot) & = [f_1(\cdot),...,f_d(\cdot)]^\top \sim \prod_{j=1}^d \mathcal{GP}(b_j(\cdot), k^m(\cdot,\cdot)), \\
{\bf h}(\cdot) & = [h_1(\cdot),...,h_d(\cdot)]^\top \sim \prod_{j=1}^d \mathcal{GP}(c_j(\cdot), k^s(\cdot,\cdot)).
\label{eq:gp_prior}
\end{align}
%%%%
Here ${\bf b}(\cdot) = [b_1(\cdot),...,b_d(\cdot)]^\top$,  ${\bf c}(\cdot) = [c_1(\cdot),...,c_d(\cdot)]^\top$ are the GP mean functions which can be modeled by deep neural networks, and the GP covariance functions of ${\bf f}$ and ${\bf h}$ are denoted by $k^m$ and $k^s$, respectively, where we share the same covariance function across dimensions for simplicity. 

%In summary, this model basically can take into account the uncertainty of both mean and covariance functions for the encoder network in VAE. 

\textbf{Relation to the SAVI}.  
Note that the GP-priored variational density model in  (\ref{eq:q_z_xfh}--\ref{eq:gp_prior}) can be equivalently written as:
%%%%
\begin{equation}
q({\bf z}|{\bf x},{\bf f},{\bf h}) = \mathcal{N}\big( %{\bf z};
{\bf b}({\bf x}) + {\bf f}({\bf x}), \textrm{Diag}({\bf c}({\bf x}) + {\bf h}({\bf x}))^2\big), 
\label{eq:q_z_xfh_alt}
\end{equation}
%%%%
where ${\bf f}$ and ${\bf h}$ now follow {\em zero-mean} Gaussian processes. % with the covariances $k^m$ and $k^s$, respectively. %(See Sec.~\ref{sec:bias_interpretation} for the interpretation that can be viewed from this form). 
%If we regard ${\bf b}({\bf x})$ and ${\bf c}({\bf x})$ as the (deterministic) mean and standard deviation functions of the conventional VAE, 
If we view the VAE's point estimate inference model as: $q({\bf z}|{\bf x}) = \mathcal{N}\big( %{\bf z}; 
{\bf b}({\bf x}), \textrm{Diag}({\bf c}({\bf x}))^2\big)$, 
then (\ref{eq:q_z_xfh_alt}) effectively models the {\em discrepancy} between the VAE’s %variational posterior 
$q({\bf z}|{\bf x})$ and the true posterior $p_{\bm{\theta}}({\bf z}|{\bf x})$ via stochastic noise models. Recall that in order to reduce this discrepancy, the semi-amortized approaches perform extra SVI gradient updates starting from ${\bf b}({\bf x})$ and ${\bf c}({\bf x})$ at test time on the fly. Instead, we aim to learn the discrepancy using Bayesian neural networks ${\bf f}$ and ${\bf h}$ (GP as a special case; see Sec.~\ref{sec:lik_posterior}), resulting in a faster and more accurate amortized inference model by taking into account uncertainty in a principled manner. 
%This is related (similar) to another version of MAML where the displacement is modeled/learned by another neural network (instead of gradient update on the fly). 

For instance, the GP posterior $p({\bf f},{\bf h}|\mathcal{D})$ can predict the above-mentioned discrepancy accurately, while their variances (e.g., $\mathbb{V}(f({\bf x})|\mathcal{D})$) %and $\mathbb{V}(f({\bf x})|\mathcal{D})$ 
%allow us to 
can serve as {\em gauge} that quantifies the degree of (instance-wise) uncertainty/difficulty in posterior approximation via the amortized inference network. To this end, we describe a reasonable likelihood model to establish a GP framework, and derive an efficient GP posterior inference algorithm in what follows. 

%\textbf{Motivation/Intuition}. According to the recent works on amortization error and semi-amortized VAE, such an encoder is not accurate enough, and we need to consider displacement (or error or residual) between $\mathcal{N}\big({\bf z}; {\bf b}({\bf x}), \textrm{Diag}({\bf c}({\bf x}))^2\big)$ and the true posterior $p_{\bm{\theta}}({\bf z}|{\bf x})$. Instead of online gradient update methods (eg, SAVAE), we regard this error/residual as a {\em stochastic noise} in the encoder distribution. Specifically, we let ${\bf f}({\bf x})$ and ${\bf h}({\bf x})$ responsible for ``residual or compensation of ${\bf b}({\bf x})$ and ${\bf c}({\bf x})$ for the mismatch'', and by placing prior on ${\bf f}({\bf x})$ and ${\bf h}({\bf x})$, perhaps the posterior ${\bf f},{\bf h}|\mathcal{D}$ may give us correct displacement to be refined for $\mathcal{N}\big({\bf z}; {\bf b}({\bf x}), \textrm{Diag}({\bf c}({\bf x}))^2\big)$. 

%%%%%%%%%%%%%%%%%%%%%%%%%%%%%%%%%%%%%%%%%%%%%%%%%%%%%%%%%%%%%%%%%%%%%%%%%%%%%%%
\subsection{Likelihood Model and GP Posterior Inference}\label{sec:lik_posterior}

To establish a valid Bayesian framework, we define a likelihood model, that is, the compatibility score of how each individual instance ${\bf x} \sim \mathcal{D}$ is likely to be generated under the given functions ${\bf f}$ and ${\bf h}$. A reasonable choice is the variational lower bound (\ref{eq:vae_logpx_elbo}), which we denote as:
%%%%
\begin{equation}
\mathcal{L}_{\bm{\theta}}({\bf f}, {\bf h}; {\bf x}) := \mathbb{E}_{q({\bf z}|{\bf x},{\bf f},{\bf h})} \big[ \log p_{\bm{\theta}}({\bf x},{\bf z}) - \log q({\bf z}|{\bf x},{\bf f},{\bf h}) \big].
\label{eq:logpx_elbo}
\end{equation}
%%%%
%of $\log p_{\bm{\theta}}({\bf x})$, namely 
Clearly $\log p_{\bm{\theta}}({\bf x}) \geq 
\mathcal{L}_{\bm{\theta}}({\bf f}, {\bf h}; {\bf x})$, and (\ref{eq:logpx_elbo}) can serve as surrogate\footnote{Technically, $e^{\mathcal{L}}$ may not be a valid density (integration not equal to $1$), and one has to deal with the difficult normalizing partition function in principle. For simplicity, we do not consider it and regard $\mathcal{L}$ as {\em unnormalized} log-likelihood function.} for the log-likelihood function $\log p({\bf x}|{\bf f},{\bf h})$. 
Given the data $\mathcal{D} =\{{\bf x}^i\}_{i=1}^N$, combining the GP priors and the likelihood model leads to the GP posterior, %$p({\bf x}|{\bf f},{\bf h}) := e^{\mathcal{L}_{\bm{\theta}}({\bf f}, {\bf h}; {\bf x})}$, the GP  posterior inference given the data $\mathcal{D}\ni {\bf x}$ is to compute:
%%%%
\begin{equation}
%\log p({\bf f},{\bf h}|\mathcal{D}) =_c \log p({\bf f}) + \log p({\bf h}) + \sum_{{\bf x}\in\mathcal{D}} \mathcal{L}_{\bm{\theta}}({\bf f}, {\bf h}; {\bf x}),
p({\bf f},{\bf h}|\mathcal{D}) \propto p({\bf f}) \ p({\bf h}) \ \prod_{{\bf x} \in \mathcal{D}}  %\prod_{i=1}^N 
\exp \big( \mathcal{L}_{\bm{\theta}}({\bf f}, {\bf h}; {\bf x}) \big).
\label{eq:gp_posterior}
\end{equation}
%%%%
%where $=_c$ refers to equality up to constant. 
However, solving (\ref{eq:gp_posterior}) requires time and memory cubic in the number of data points $N$, which is prohibitive for large-scale data. Although there exist efficient scalable approximate inference techniques in the GP literature~\cite{quinonero05,snelson06,titsias09,dezfouli15,vff_gp}, here we adopt the {\em linear deep kernel} trick~\cite{deep_kernel,dkl16}, %The key idea is to  A more detailed description can also be found in the Supplement.
which we briefly summarize below.
%can be viewed as a special case of Bayesian neural networks. 

%%%%%%%%%%%
%\subsubsection{GP Approximation via Deep Kernel}\label{sec:deep_kernel}

\textbf{Linear deep kernel trick for approximating GP.} %\hl{(REPHRASE BELOW)} 
A random (scalar) function $f({\bf x})$ that follows the 0-mean GP with covariance (kernel) $k$, namely $f(\cdot) \sim \mathcal{GP}(0, k(\cdot,\cdot))$, can be represented as a linear form with an explicit feature space mapping. Consider a feature mapping $\bm{\psi}:\mathcal{X} \to \mathbb{R}^p$ such that the covariance function is approximated as inner product in the feature space (of dimension $p$), namely $k({\bf x},{\bf x}') \approx \bm{\psi}({\bf x})^\top \bm{\psi}({\bf x}')$. Now, introducing the $p$-variate random vector ${\bf w} \sim \mathcal{N}({\bf 0},{\bf I})$, allows us to write the GP function as $f({\bf x}) = {\bf w}^\top {\boldsymbol \psi}({\bf x})$. It is because $\textrm{Cov}(f({\bf x}),f({\bf x}')) = \textrm{Cov}({\bf w}^\top \bm{\psi}({\bf x}), {\bf w}^\top  \bm{\psi}({\bf x}')) =  \bm{\psi}({\bf x})^\top \bm{\psi}({\bf x}') \approx k({\bf x},{\bf x}')$. 
A main advantage of this representation is that we can turn the non-parametric GP into a parametric Bayesian model, where the posterior inference can be done on the finite dimensional random vector ${\bf w}$ instead. %More specifically, 
%%%%
% \begin{equation}
% p({\bf W}|\mathcal{D}) \propto \mathcal{N}({\bf w}; {\bf 0}, {\bf I}) \cdot \prod_{{\bf x}\in\mathcal{D}} l({\bf w}^\top {\boldsymbol \phi}({\bf x})), 
% \label{eq:w_posterior}
% \end{equation}
%%%%
%where $l({\bf w}^\top {\boldsymbol \phi}({\bf x}))$ is the likelihood function for input ${\bf x}$ which depends on $f$ only through $f({\bf x}) = {\bf w}^\top {\boldsymbol \phi}({\bf x})$. 
The feature mapping $\bm{\psi}(\cdot)$ can be modeled as a deep neural network, and its weight parameters constitute the covariance (kernel) parameters of the GP. This way, we can (approximately) view GP as a special case of Bayesian neural networks where we treat the final fully connected layer ${\bf w}$ as random~\cite{gp_bnn_book,gp_bnn_dnngp,gp_bnn_gpwdnn,gp_bnn_dcngp}. 
Note that although this is rather a simplified form of the deep kernel~\cite{dkl16} by applying the linear kernel on the outputs of $\bm{\psi}(\cdot)$, it has been widely used with great success~\citep{deep_kernel,frcl20}. 

Returning to our GP posterior inference (\ref{eq:gp_posterior}), the two GP-priored functions can be written as: $f_j({\bf x}) = {\bf w}_j^\top \bm{\psi}^m({\bf x})$ and $h_j({\bf x}) = {\bf u}_j^\top \bm{\psi}^s({\bf x})$ for $j=1,\dots,d$, where ${\bf w}_j$'s and ${\bf u}_j$'s are mutually independent $p$-variate random vectors from $\mathcal{N}({\bf 0},{\bf I})$. The feature functions $\bm{\psi}^m,\bm{\psi}^s:\mathcal{X}\rightarrow \mathbb{R}^p$ are deep neural networks that define the covariance functions: $k^m({\bf x},{\bf x}') = \bm{\psi}^m({\bf x})^\top \bm{\psi}^m({\bf x}')$,  $k^s({\bf x},{\bf x}') = \bm{\psi}^s({\bf x})^\top \bm{\psi}^s({\bf x}')$. By letting ${\bf W} = [{\bf w}_1,\dots,{\bf w}_d]^\top$ and ${\bf U} = [{\bf u}_1,\dots,{\bf u}_d]^\top$ be the $(d \times p)$ matrices with the random vectors in the rows, we have %more succinct representation,
${\bf f}({\bf x}) = [f_1({\bf x}),...,f_d({\bf x})]^\top = {\bf W} \bm{\psi}^m({\bf x})$, % and 
${\bf h}({\bf x}) = [h_1(\cdot),...,h_d(\cdot)]^\top = {\bf U} \bm{\psi}^s({\bf x})$. The inference in (\ref{eq:q_z_xfh_alt}) can be written as $q({\bf z}|{\bf x},{\bf W},{\bf U})$ that equals:
%in terms of ${\bf W}$, ${\bf U}$, %namely 
%%%%
%\vspace{-1.2em}
\begin{equation}
%q({\bf z}|{\bf x},{\bf W},{\bf U}) =
\mathcal{N}\big( {\bf b}({\bf x}) + {\bf W} \bm{\psi}^m({\bf x}), \textrm{Diag}({\bf c}({\bf x}) + {\bf U} \bm{\psi}^s({\bf x}))^2 \big), 
\label{eq:q_z_xWU}
\end{equation}
%%%%
%\footnote{(We add small $\sigma_{min}^2$ to the diagonals of the covariance matrix to avoid the numerical issue.)}
while (\ref{eq:gp_posterior}) becomes:
%%%%
\begin{align}
& p({\bf W},{\bf U}|\mathcal{D}) \propto \mathcal{N}({\bf W}; {\bf 0}, {\bf I}) \mathcal{N}({\bf U}; {\bf 0}, {\bf I}) \prod_{{\bf x}\in\mathcal{D}} e^{\mathcal{L}_{\bm{\theta}}({\bf W}, {\bf U}; {\bf x})}, \nonumber \\
& \ \ \ \ \textrm{where} \ \mathcal{L}_{\bm{\theta}}({\bf W}, {\bf U}; {\bf x}) := \mathbb{E}_{q} \bigg[ \log \frac{ p_{\bm{\theta}}({\bf x},{\bf z}) }{ q({\bf z}|{\bf x},{\bf W},{\bf U}) } \bigg].
\label{eq:WU_posterior}
\end{align}
%%%%
%where $\mathcal{L}_{\bm{\theta}}({\bf W}, {\bf U}; {\bf x}) := \mathbb{E}_{q} \big[ \log \frac{ p_{\bm{\theta}}({\bf x},{\bf z}) }{ q({\bf z}|{\bf x},{\bf W},{\bf U}) } \big]$. 
%Computing $p({\bf W},{\bf U}|\mathcal{D})$ % of (\ref{eq:WU_posterior})
%Since (\ref{eq:WU_posterior}) is intractable, we
We approximate (\ref{eq:WU_posterior}) %with the variational density 
by $q({\bf W},{\bf U})$ %to approximate it. %the posterior. 
%We assume a %fully
%factorized Gaussian
defined as:%\footnote{We let $\bm{\Sigma}_j$ and $\bm{\Gamma}_j$ be {\em full} covariances. This would not incur overfitting since we only enrich the capacity of the variational density $q({\bf W},{\bf U})$ as discussed in~\cite{titsias09}.} %that is, 
%%%%
\begin{equation}
q({\bf W},{\bf U};\bm{\Lambda}) = %q({\bf W}) q({\bf U}) =
  %\mathcal{N}({\bf W}; \bm{\mu}, \bm{\Sigma}) 
  %\mathcal{N}({\bf U}; \bm{\eta}, \bm{\Gamma}),
  \prod_{j=1}^d \mathcal{N}({\bf w}_j; \bm{\mu}_j, \bm{\Sigma}_j) \ \mathcal{N}({\bf u}_j; \bm{\eta}_j, \bm{\Gamma}_j) 
\label{eq:varinf_q}
\end{equation}
%%%% 
where $\bm{\Lambda} := \{ \bm{\mu}_j, \bm{\Sigma}_j, \bm{\eta}_j, \bm{\Gamma}_j \}_{j=1}^d$ constitutes the GP  variational parameters.
%where $\bm{\Sigma}_j$ and $\bm{\Gamma}_j$ are diagonal matrices. %, and $\{ \bm{\mu}, \bm{\Sigma}, \bm{\eta}, \bm{\Gamma} \}$ constitute the variational parameters for the GP.

%%%%%%%%%%%%%%%%%%%%%%%%%%%%%%%%%%%%%%%%%%%%%%%%%%%%%%%%%%%%%%%%%%%%%%%%%%%%%%%
\subsection{GP Posterior Marginalized Encoder}\label{sec:gp_encoder}

%%%%%%
%\textbf{GP posterior marginalized encoder.} 
Before we proceed to GP inference and learning (Sec.~\ref{sec:gp_learning}), we derive the posterior averaged encoder distribution, %more specifically 
%%%%
\begin{equation}
q({\bf z}|{\bf x}) = \iint q({\bf W},{\bf U}) q({\bf z}|{\bf x},{\bf W},{\bf U}) d {\bf W} d{\bf U}.
\label{eq:qzx}
\end{equation}
%%%%
Note that (\ref{eq:qzx}) can be seen as the final latent inference model of our GP VAE model, where the uncertainty captured in the GP posterior $q({\bf W},{\bf U})$ is all marginalized out. For instance, the test log-likelihood score $\log p({\bf x})$ under our model can be estimated by the importance weighted sampling method~\cite{iwae} %using (\ref{eq:qzx}) 
as the proposal distribution. As it also appears in the GP learning in the next section, we provide the  derivation for $q({\bf z}|{\bf x})$ here. 

%Once the learning is done, that is, once we have the posterior approximate $q({\bf W},{\bf U})$, the encoder can be formed by marginalizing out ${\bf W}$ and ${\bf U}$ wrt the posterior distribution. That is,

Although the two terms in the integrand of (\ref{eq:qzx}) are both Gaussians, it is infeasible to have a close-form formula due to the dependency of the covariance of (\ref{eq:q_z_xWU}) on ${\bf U}$. % for the integration. 
Instead, our approximation strategy is to view (\ref{eq:qzx}) as (a limit of) a mixture of Gaussians, $\sum_i \alpha_i \mathcal{N}({\bf z}; {\bf m}_i, {\bf V}_i)$ where $\alpha_i$ corresponds to $q({\bf W},{\bf U})$ with index $i$ mapped to $({\bf W},{\bf U})$, and ${\bf m}_i, {\bf V}_i$ denote the mean and covariance of $q({\bf z}|{\bf x},{\bf W},{\bf U})$ in (\ref{eq:q_z_xWU}). Since a Gaussian mixture %of Gaussians 
can be approximated by a single Gaussian by the second-order moment matching\footnote{Equivalent to $\min_{\{{\bf m}, {\bf V}\}} \textrm{KL}\big( \sum_i \alpha_i \mathcal{N}({\bf m}_i, {\bf V}_i) || \mathcal{N}({\bf m}, {\bf V}) \big)$.}, namely $\sum_i \alpha_i \mathcal{N}({\bf m}_i, {\bf V}_i) \approx \mathcal{N}({\bf m}, {\bf V})$ where ${\bf m} = \sum_i \alpha_i {\bf m}_i$ and ${\bf V} = \sum_i \alpha_i ({\bf m}_i {\bf m}_i^\top +{\bf V}_i) - {\bf m}{\bf m}^\top$, applying it to (\ref{eq:qzx}) yields: $q({\bf z}|{\bf x}) \approx \mathcal{N}({\bf z}; {\bf m}({\bf x}), \textrm{Diag}({\bf v}({\bf x})))$ where
%%%%
\begin{align}
%q({\bf z}|{\bf x}) \approx \prod_{j=1}^d \mathcal{N}(z_j; m_j, V_j),
m_j({\bf x}) &= b_j({\bf x}) + \bm{\mu}_j^\top \bm{\psi}^m({\bf x}), \nonumber \\
v_j({\bf x}) &= c_j({\bf x})^2 + 2\bm{\eta}_j^\top \bm{\psi}^s({\bf x}) c_j({\bf x}) + \bm{\psi}^m({\bf x})^\top \bm{\Sigma}_j \bm{\psi}^m({\bf x}) %\nonumber \\ 
%& \ \ \ \ \ \ \ 
+ \bm{\psi}^s({\bf x})^\top \big( \bm{\eta}_j \bm{\eta}_j^\top + \bm{\Gamma}_j \big) \bm{\psi}^s({\bf x}), \label{eq:qzx_mm}
\end{align}
%%%%
for $j=1,\dots,d$. %See Supplement for detailed derivations. 

Note from (\ref{eq:qzx_mm}) that as a special case, $q({\bf z}|{\bf x})$ becomes the standard VAE's encoder distribution with means $b_j({\bf x})$ and variances $c_j({\bf x})^2$ if the GP posterior is ignored (i.e., $\bm{\Lambda}=\{ \bm{\mu}_j, \bm{\Sigma}_j, \bm{\eta}_j, \bm{\Gamma}_j \}={\bf 0}$ to lead to the deterministic zero noise model). %$\bm{\mu}_j=\bm{\Sigma}_j=\bm{\eta}_j=\bm{\Gamma}_j={\bf 0}$).
And our learned GP posterior (non-zero $\bm{\Lambda}$) informs us how the deviation from the true posterior $p_{\bm{\theta}}({\bf z}|{\bf x})$ can be determined and compensated, namely by (\ref{eq:qzx_mm}).

%%%%%%%%%%%%%%%%%%%%%%%%%%%%%%%%%%%%%%%%%%%%%%%%%%%%%%%%%%%%%%%%%%%%%%%%%%%%%%%
\subsection{GP Inference and Learning}\label{sec:gp_learning}

%%%%%%
%\textbf{GP ELBO objective.} 

Now we describe how the variational GP inference (i.e., optimizing $\bm{\Lambda}$ in $q({\bf W},{\bf U};\bm{\Lambda})$) can be done. Similar to other GP variational leanring, the objective function that we will derive establishes a lower bound of the model's data likelihood, and hence we can learn the {\em model parameters} as well by maximizing the lower bound (empirical Bayes). The model parameters consist of the parameters in the GP mean and covariance functions (i.e., the weight parameters of the deep networks ${\bf b}({\bf x})$, ${\bf c}({\bf x})$, $\bm{\psi}^m({\bf x})$, and $\bm{\psi}^s({\bf x})$), and those in the likelihood model (i.e., $\bm{\theta}$ in the decoder $p_{\bm{\theta}}({\bf x}|{\bf z})$).

To approximate $q({\bf W},{\bf U}) \approx p({\bf W},{\bf U}|\mathcal{D})$, we aim to minimize %their KL divergence 
$\textrm{KL} \big( q({\bf W},{\bf U}) || p({\bf W},{\bf U}|\mathcal{D}) \big)$, and it can be shown that the KL can be written as follows (Supplement for details): 
%%%%
\vspace{-0.3em}
\begin{equation}
\vspace{-0.3em}
\textrm{KL}(q||p) = \log \hat{p}_{\bm{\theta}}(\mathcal{D}) - \sum_{{\bf x} \in\mathcal{D}} \textrm{ELBO}(\bm{\theta}, \bm{\Lambda}; {\bf x}),
\label{eq:kl_elbo}
\end{equation}
%%%%
where $\hat{p}_{\bm{\theta}}(\mathcal{D}) = \mathbb{E}_{{\bf W},{\bf U} \sim \mathcal{N}({\bf 0}, {\bf I})} \big[ \prod_{{\bf x}} e^{\mathcal{L}_{\bm{\theta}}({\bf W}, {\bf U}; {\bf x})} \big]$ is the marginal data likelihood using our surrogate %likelihood function 
$\mathcal{L}$ in (\ref{eq:WU_posterior}), %$\bm{\Lambda} := \{ \bm{\mu}_j, \bm{\Sigma}_j, \bm{\eta}_j, \bm{\Gamma}_j \}_{j=1}^d$ constitutes the variational parameters for the GP, and finally 
and %$\textrm{ELBO}(\bm{\theta}, \bm{\Lambda})$ is defined as:
%%%%
\begin{align}
\textrm{ELBO} %(\bm{\theta}, \bm{\Lambda}; {\bf x}) \ 
:= - \mathbb{E}_{q({\bf W},{\bf U})}[ \textrm{KL}( q({\bf z}|{\bf x},{\bf W},{\bf U}) || p({\bf z}) ) ] \ + \ 
%\mathbb{E}_{q({\bf W},{\bf U}) q({\bf z}|{\bf x},{\bf W},{\bf U})} 
\mathbb{E}_{q({\bf z}|{\bf x})} \big[ \log p_{\bm{\theta}}({\bf x}|{\bf z}) \big] \ - \ %\label{eq:elbo_1} %\\
%& \ \ \ \ \ \ 
\frac{1}{N} \textrm{KL}( q({\bf W},{\bf U}) || \mathcal{N}({\bf 0}, {\bf I}) ). %\nonumber %\label{eq:elbo_3} 
\label{eq:elbo}
\end{align}
%%%%
%The ELBO is a lower bound of the (surrogate) data log-likelihood $\log \hat{p}_{\bm{\theta}}(\mathcal{D})$, as evident from (\ref{eq:kl_elbo}). Thus optimizing the ELBO with respect to the variational parameters $\bm{\Lambda}$ tightens the bound, while maximizing it with respect to the model parameters leads to optimal data fitting. 
%
We now discuss how individual terms in the ELBO (\ref{eq:elbo}) can be derived. The last term of (\ref{eq:elbo}) is the KL divergence between Gaussian densities, and admits a close form. The second term is the expected log-likelihood with respect to the GP posterior marginalized encoder $q({\bf z}|{\bf x})$, (\ref{eq:qzx_mm}), %in Sec.~\ref{sec:gp_encoder}, 
and we can do this by Monte Carlo estimation with the well-known reparametrization trick~\cite{vae14}.  %using the moment-matching approximation (\ref{eq:qzx_mm}). 
%That is, we have the reparametrized sample ${\bf z}\sim q({\bf z}|{\bf x})$ where 
%More specifically, 
% That is, we take the sample ${\bf z}\sim q({\bf z}|{\bf x})$ as, for $j=1,\dots,d$,
% %%%%
% \begin{equation}
% z_j = m_j({\bf x}) + \epsilon_j v_j({\bf x}), \ \ \ \ \epsilon_j \sim \mathcal{N}(0,1).
% \end{equation}
% %%%%
Finally, the first term in (\ref{eq:elbo}) is the Gaussian averaged KL divergence between Gaussians, and thus it can also admit a closed form. More specifically, it equals: % (Supplement for details):
%%%%
%\vspace{-0.3em}
\begin{align}
\frac{1}{2} \sum_{j=1}^d \bigg( 
  v_j({\bf x}) + 
  \big( b_j({\bf x}) + \bm{\mu}_j^\top \bm{\psi}^m({\bf x}) \big)^2 - 1 \ - \
%& \ \ \ \ \ \ \ \ \ \ \ \ \ \ 
\mathbb{E}_{\mathcal{N}({\bf u}_j; \bm{\eta}_j, \bm{\Gamma}_j)}\Big[\log \big( c_j({\bf x}) + {\bf u}_j^\top \bm{\psi}^s({\bf x}) \big)^2 \Big]
%
%m_j({\bf x}) &= b_j({\bf x}) + \bm{\mu}_j^\top \bm{\psi}^m({\bf x}), \nonumber \\
%v_j({\bf x}) &= c_j({\bf x})^2 + 2\bm{\eta}_j^\top \bm{\psi}^s({\bf x}) c_j({\bf x}) + \bm{\psi}^m({\bf x})^\top \bm{\Sigma}_j \bm{\psi}^m({\bf x}) \nonumber \\ 
%& \ \ \ \ \ \ \ \ + \bm{\psi}^s({\bf x})^\top \big( \bm{\eta}_j \bm{\eta}_j^\top + \bm{\Gamma}_j \big) \bm{\psi}^s({\bf x}), 
\bigg). %\nonumber
%\vspace{-1.2em}
\label{eq:ekl}
\end{align}
%%%%
The last term in (\ref{eq:ekl}) is essentially a Gaussian expected squared $\log$ function, which can be written as a closed form, albeit complicated, using the confluent hyper-geometric function~\cite{lloyd}. However, for simplicity % and to avoid the overhead of building an accurate multi-resolution look-up table, 
we estimate it using the reparametrized Monte-Carlo method.  %similarly as before. 

%%%%%%%%
\textbf{Summary.} % of the learning algorithm.} 
%We summarize 
The overall learning steps are as follows: %steps to learn our GP VAE model: %is as follows:
%%%%
\vspace{-0.8em}
\begin{enumerate}
\item Initialize the variational parameters $\bm{\Lambda}$ and the model parameters $\bm{\theta}$, ${\bf b}({\bf x})$, ${\bf c}({\bf x})$, $\bm{\psi}^m({\bf x})$, and $\bm{\psi}^s({\bf x})$.
\vspace{-0.5em}
\item Repeat until convergence:
\begin{enumerate}
    \vspace{-0.2em}
    \item Estimate %the GP posterior marginalized encoder 
    the marginalized $q({\bf z}|{\bf x})$ using (\ref{eq:qzx_mm}).
    \item Optimize the ELBO (\ref{eq:elbo}) %with respect to 
    wrt all parameters. %\footnote{Alternatively, we can partition the parameters into several groups (e.g., $\bm{\Lambda}$ vs.~the rest), and do block-coordinate optimization.}.
\end{enumerate}
\vspace{-0.5em}
\item (At test time) The GP marginalized encoder $q({\bf z}|{\bf x})$ can be used to perform reconstruction, and evaluate the test likelihood $p_{\bm{\theta}}({\bf x})$, e.g., using the importance weighted sampling method~\cite{iwae}. The uncertainty (variance) of the posterior noise $\textrm{Tr}\mathbb{V}({\bf f}({\bf x})|\mathcal{D})$ (similarly for ${\bf h}$) can be approximately estimated as $\bm{\psi}^m({\bf x})^\top \big(  \sum_{j=1}^d \bm{\Sigma}_j \big) \bm{\psi}^m({\bf x})$. 
\end{enumerate} 
%%%%

%%%%%%%%%%%%%%%%%%%%%%%%%%%%%%%%%%%%%%%%%%%%%%%%%%%%%%%%%%%%%%%%%%%%%%%%%%%%%%%
%%%%%%%%%%%%%%%%%%%%%%%%%%%%%%%%%%%%%%%%%%%%%%%%%%%%%%%%%%%%%%%%%%%%%%%%%%%%%%%
\section{Related Work}\label{sec:related}

\begin{comment}
 % beyond the axis-parallel diagonal covariances in the VAE. 
%
%\textbf{Functional gradient.} 
Our use of functional gradient in designing a learning objective stems from the framework in~\cite{funcgrad_friedman,funcgrad_mason}. Mathematically elegant and flexible in the learning criteria, the framework was more recently exploited in~\cite{pfg} to unify seemingly different machine learning paradigms. %, including variational inference, adversarial learning, and reinforcement learning. 
%
%\textbf{Other mixture modeling.} 
%There were 
Several mixture-based approaches aimed to extend the representational capacity of the variational inference model. In~\cite{sivi} the variational parameters were mixed with a flexible distribution. %In~\cite{zobay} Gaussian-mixture approximation is used, while 
In~\cite{vampprior} the prior is modeled as a mixture (aggregate posterior). 
%Another attempt to enrich the VAE's model complexity is the so called 
%%%%%\begin{comment}
The hierarchical decompositional mixture~\cite{spvae} %. Although the model is termed a mixture, it 
is a hybrid model of VAE and the sum-product network~\cite{spn} (VAEs as leaf nodes), %leading to the model family 
very distinct from the framework that we focus on in this paper. %; hence, empirical comparison to this model is out of scope. 
%%%%%\end{comment}
\end{comment}

As enumerating all related literature in this section can be infeasible, we briefly review some of the recent works that are highly related with ours. The issue of amortization error in VAE was raised in~\cite{cremer18}, after which several semi-amortized approaches have been attempted~\cite{savae_ykim,savae_marino,savae_krishnan} that essentially follow a few SVI gradient steps at test time. An alternative line of research approaches the problem by enlarging the representational capacity of the encoder network, including the flow-based models that apply nonlinear invertible transformations to VAE's variational posterior~\cite{hfvae,iafvae}. 
Recently~\citep{mkim2020neurips} proposed a greedy recursive mixture estimation method for the encoder in VAE, where the idea is to  iteratively augment the current mixture with new components to maximally reduce the divergence between the variational and the true posteriors. 

In parallel, there have been previous attempts to apply the Bayesian approach to the VAE modeling. However, they are in nature different from our random function modeling of the encoder uncertainty. 
The Bayesian Variational VAE~\citep{bv_vae} rather focused on modeling uncertainty in the {\em decoder} model, %whereas ours aims at capturing uncertainty in the inference network. Moreover, 
and their main focus is how to deal with out-of-distribution samples in the test set,  hence more aligned with transfer learning.
The Compound VAE~\citep{compound_vae} also tackled the similar problem of reducing the amortization gap of the VAE, however, their variational density modeling is less intuitive, inferring the latent vector ${\bf z}$ and the encoder weights ${\bf W}$ from each data instance. Note that we have more intuitive Bayesian inference for the encoder parameters,  $q({\bf W}|\mathcal{D})$ given the {\em entire} training data $\mathcal{D}$. Their treatment is deemed to augment the latent ${\bf z}$ with the weights ${\bf W}$ in the conventional VAE.
The Variational GP~\citep{vgp}, although looking similar to ours, is not specifically aimed for the VAE and amortized inference, but for general Bayesian inference. In turn, they built the posterior model using a GP function defined on the Gaussian distributed {\em latent input} space, instead of defining GP on the input data as we did.

%%%%%%%%%%%%%%%%%%%%%%%%%%%%%%%%%%%%%%%%%%%%%%%%%%%%%%%%%%%%%%%%%%%%%%%%%%%%%%%
%%%%%%%%%%%%%%%%%%%%%%%%%%%%%%%%%%%%%%%%%%%%%%%%%%%%%%%%%%%%%%%%%%%%%%%%%%%%%%%
\section{Evaluations}\label{sec:expmt}

We evaluate our Gaussian process VAE model on several benchmark datasets to show its improved performance over the existing state-of-the-arts. Our focus is two-fold: 1) improved test likelihood scores, and 2) faster test time inference than semi-amortized methods. We also contrast with the flow-based models that employ high capacity encoder networks. 
The competing approaches are as follows:
%%%%
\begin{itemize}
\vspace{-0.6em}
\item \textbf{VAE}: The standard VAE model with amortized inference~\cite{vae14,vae14r}.
\item \textbf{SA}: The semi-amortized VAE~\cite{savae_ykim}. We fix the SVI gradient step size as $10^{-3}$, but vary the number of SVI steps from $\{1, 2, 4, 8\}$. 
\item \textbf{IAF}: The autoregressive-based flow model for the encoder $q({\bf z}|{\bf x})$~\cite{iafvae}, which has richer expressive capability than the VAE's post-Gaussian encoder. % adopted in VAE models. 
The number of flows %(i.e., the number of compositions) 
is chosen from $\{1, 2, 4, 8\}$. 
\item \textbf{HF}: The Householder flow encoder model that represents the full covariance using the Householder transformation~\cite{hfvae}. The number of flows %(i.e., the number of compositions) 
is chosen from $\{1, 2, 4, 8\}$. 
\item \textbf{ME}: To enlarge the representational capacity of the encoder network, another possible baseline is a mixture model. More specifically, the inference model is defined as: $q({\bf z}|{\bf x}) = \sum_{m=1}^M \alpha(m|{\bf x}) q_m({\bf z}|{\bf x})$, where $q_m({\bf z}|{\bf x})$ are amortized inference models (e.g., having the same network architectures as the VAE's encoder network), and  $\alpha(m|{\bf x})$ are mixing proportions, dependent on the input ${\bf x}$, which can be modeled by a single neural network. The mixture encoder (ME) model is trained by gradient ascent to maximize the lower bound of $\log p({\bf x})$ similarly as the VAE. The number of mixture components $M$ is chosen from $\{1,2,4,8\}$.
\item \textbf{RME}: The recursive mixture estimation method for the encoder in VAE~\citep{mkim2020neurips}, which showed superiority to ME's blind mixture estimation. 
\item \textbf{GPVAE}: Our proposed GP encoder model. The GP means and  feature functions have the same network architectures as the VAE's encoder. 

%The GP means of mean and stdev of the variational posterior are initialized with mean and stdev of the VAE's trained encoder (and updated thereafter). For the details of the hyperparameters, refer to the Supplement. 
\end{itemize}
%%%%

%%%%%%%%

\textbf{Datasets.} We use the following five benchmark datasets: \textbf{MNIST}~\cite{mnist}, %$(28 \times 28 \times 1)$. 
\textbf{OMNIGLOT}~\cite{omniglot}, %: $24,345$ training images and $8,070$ test images where each image is of dimension $(28 \times 28 \times 1)$. We randomly hold out $10\%$ of the training set as a validation set. 
    %\item \textbf{FashionMNIST}~\cite{fashion_mnist}: $60,000$ training images and $10,000$ test images where each image is of dimension $(28 \times 28 \times 1)$. We randomly hold out $10\%$ of the training set as a validation set. 
\textbf{CIFAR10}\footnote{Results on CIFAR10 can be found in the Supplement.
}~\cite{cifar10}, %: $50,000$ training images and $10,000$ test images where each image is of dimension $(32 \times 32 \times 3)$. We randomly hold out $10\%$ of the training set as a validation set. 
\textbf{SVHN}~\cite{svhn}, % $73,257$ training images and $26,032$ test images where each image is of dimension $(32 \times 32 \times 3)$. We randomly hold out $10\%$ of the training set as a validation set. 
and \textbf{CelebA}~\cite{celeba}. %: $202,599$ tightly cropped face images of size $(64 \times 64 \times 3)$. 
For CelebA, we use tightly cropped face images of size $(64 \times 64 \times 3)$, and randomly split the data into $80\%/10\%/10\%$ train/validation/test sets. For the other datasets, we follow the %train/test 
partitions provided in the data, with $10\%$ of the training sets randomly held out for validation. 

\definecolor{lor}{rgb}{1,0.85,0}
\definecolor{or}{rgb}{1,0.60,0}
\definecolor{dor}{rgb}{1,0.20,0}
\newcommand\Tstrut{\rule{0pt}{2.2ex}}         % = `top' strut
\newcommand\Bstrut{\rule[-0.9ex]{0pt}{0pt}}   % = `bottom' strut

%%%% 
\begin{table}[t]
\vspace{-0.6em}
\centering
\caption{(MNIST) 
Test log-likelihood scores (unit in nat).  %estimated by the importance weighted sampling~\cite{iwae}. % with 100 samples. 
The figures in the parentheses next to model names indicate: the number of SVI steps in SA,  the number of flows in IAF and HF, and the number of mixture components in ME and RME. The superscripts are the standard deviations. 
The best (on average) results are boldfaced in $\color{red} \textrm{\textbf{red}}$. In each column, we perform the two-sided $t$-test to measure the statistical significance of the difference between the best model (red) and each competing method. We depict those with $p$-values greater than $0.01$ as boldfaced $\color{blue} \textrm{blue}$ (little evidence of difference). So, anything plain non-colored indicates $p\leq 0.01$ (significantly different). 
Best viewed in color.
}
\label{tab:mnist}
%\vskip 0.05in
\vspace{0.3em}
\begin{small}
%\begin{scriptsize}
\begin{sc}
\centering
\begin{tabular}{lccc}
\toprule
 & $\textrm{dim}({\bf z})=10$ & $\textrm{dim}({\bf z})=20$ & $\textrm{dim}({\bf z})=50$ \\
\midrule
VAE & $685.1^{1.8}$ & $930.7^{3.9}$ & $1185.7^{3.9}$ \\ \hline
SA$^{(1)}$\Tstrut & $688.1^{2.7}$ & $921.2^{2.3}$ & $1172.1^{1.8}$ \\
SA$^{(2)}$ & $682.2^{1.5}$ & $932.0^{2.4}$ & $1176.3^{3.4}$ \\
SA$^{(4)}$ & $683.5^{1.5}$ & $925.5^{2.6}$ & $1171.3^{3.5}$ \\
SA$^{(8)}$ & $684.6^{1.5}$ & $928.1^{3.9}$ & $1183.2^{3.4}$ \\ \hline
IAF$^{(1)}$\Tstrut & $687.3^{1.1}$ & $934.0^{3.3}$ & $1180.6^{2.7}$ \\
IAF$^{(2)}$ & $677.7^{1.6}$ & $931.4^{3.7}$ & $1190.1^{1.9}$ \\
IAF$^{(4)}$ & $685.0^{1.5}$ & $926.3^{2.6}$ & $1178.1^{1.6}$ \\
IAF$^{(8)}$ & $689.7^{1.4}$ & $934.1^{2.4}$ & $1150.0^{2.2}$ \\ \hline
HF$^{(1)}$\Tstrut & $682.5^{1.4}$ & $917.2^{2.6}$ & $1204.3^{4.0}$ \\
HF$^{(2)}$ & $677.6^{2.2}$ & $923.9^{3.1}$ & $1191.5^{10.8}$ \\
HF$^{(4)}$ & $683.3^{2.6}$ & $927.3^{2.8}$ & $1197.2^{1.5}$ \\
HF$^{(8)}$ & $679.6^{1.5}$ & $928.5^{3.1}$ & $1184.1^{1.8}$ \\ \hline
ME$^{(2)}$\Tstrut & $685.7^{1.2}$ & $926.7^{3.0}$ & $1152.8^{1.7}$ \\
ME$^{(3)}$ & $678.5^{2.5}$ & $933.1^{4.1}$ & $1162.8^{4.7}$ \\
ME$^{(4)}$ & $680.0^{0.9}$ & $914.7^{2.3}$ & $ {1205.1}^{2.3}$ \\
ME$^{(5)}$ & $682.0^{1.7}$ & $920.6^{1.9}$ & $1198.5^{3.5}$ \\ \hline
RME$^{(2)}$\Tstrut & $\color{blue} {\bf 697.2}^{1.1}$ & $\color{blue} {\bf 943.9}^{1.6}$ & $1201.7^{0.9}$ \\
RME$^{(3)}$ & $\color{blue} {
\bf 698.2}^{1.1}$ & $\color{blue} {\bf 945.1}^{1.6}$ & $1202.4^{1.0}$ \\
RME$^{(4)}$ & $\color{blue} {\bf 699.0}^{1.0}$ & $\color{red} {\bf 945.2}^{1.6}$ & $1203.1^{1.0}$ \\
RME$^{(5)}$ & $\color{red} {\bf 699.4}^{2.1}$ & $\color{blue} {\bf 945.0}^{1.7}$ & $1203.7^{1.0}$ \\ \hline
GPVAE\Tstrut & $\color{blue} {\bf 696.5}^{1.5}$ & $\color{blue} {\bf 944.3}^{2.8}$ & $\color{red} {\bf 1212.9}^{3.2}$ \\
\bottomrule
\end{tabular}
\end{sc}
\end{small}
%\end{scriptsize}
\vskip 0.05in
\end{table}
%%%%

%%%%
\begin{table}[t]
\vspace{-0.6em}
\centering
\caption{(OMNIGLOT) Test log-likelihood scores (unit in nat). The same interpretation as \autoref{tab:mnist}. 
}
\label{tab:omniglot}
\vspace{+0.3em}
\begin{small}
\begin{sc}
\centering
\begin{tabular}{lccc}
\toprule
 & $\textrm{dim}({\bf z})=10$ & $\textrm{dim}({\bf z})=20$ & $\textrm{dim}({\bf z})=50$ \\
\midrule
VAE & $347.0^{1.7}$ & $501.6^{1.6}$ & $801.6^{4.0}$ \\ \hline
SA$^{(1)}$\Tstrut & $344.1^{1.4}$ & $499.3^{2.5}$ & $792.7^{7.9}$ \\
SA$^{(2)}$ & $349.5^{1.4}$ & $501.0^{2.7}$ & $793.1^{4.8}$ \\
SA$^{(4)}$ & $342.1^{1.0}$ & $488.2^{1.8}$ & $794.4^{1.9}$ \\
SA$^{(8)}$ & $344.8^{1.1}$ & $490.3^{2.8}$ & $799.4^{2.7}$ \\ \hline
IAF$^{(1)}$\Tstrut & $347.8^{1.6}$ & $489.9^{1.9}$ & $788.8^{4.1}$ \\
IAF$^{(2)}$ & $344.2^{1.6}$ & $494.9^{1.4}$ & $795.7^{2.7}$ \\
IAF$^{(4)}$ & $347.9^{1.9}$ & $496.0^{2.0}$ & $775.1^{2.2}$ \\
IAF$^{(8)}$ & $343.9^{1.4}$ & $498.8^{2.3}$ & $774.7^{2.9}$ \\ \hline
HF$^{(1)}$\Tstrut & $335.5^{1.2}$ & $488.6^{2.0}$ & $795.9^{3.3}$ \\
HF$^{(2)}$ & $340.6^{1.3}$ & $495.9^{1.8}$ & $784.5^{4.8}$ \\
HF$^{(4)}$ & $343.3^{1.2}$ & $487.0^{2.7}$ & $799.7^{3.2}$ \\
HF$^{(8)}$ & $343.3^{1.3}$ & $488.3^{2.4}$ & $794.6^{4.0}$ \\ \hline
ME$^{(2)}$\Tstrut & $344.2^{1.5}$ & $491.7^{1.4}$ & $793.4^{3.8}$ \\
ME$^{(3)}$ & $350.3^{1.8}$ & $491.2^{2.1}$ & $807.5^{4.9}$ \\
ME$^{(4)}$ & $337.7^{1.1}$ & $491.3^{1.8}$ & $732.0^{3.1}$ \\
ME$^{(5)}$ & $343.0^{1.4}$ & $478.0^{2.8}$ & $805.7^{3.8}$ \\ \hline
RME$^{(2)}$\Tstrut & $349.3^{1.5}$ & $508.2^{1.2}$ & $\color{blue} {\bf 821.0}^{3.1}$ \\
RME$^{(3)}$ & $349.9^{1.6}$ & $507.5^{1.1}$ & $\color{blue} {\bf 820.4}^{0.9}$ \\
RME$^{(4)}$ & $350.7^{1.7}$ & $509.0^{1.2}$ & $\color{blue} {\bf 819.9}^{0.9}$ \\
RME$^{(5)}$ & $351.1^{1.7}$ & $509.1^{1.4}$ & $\color{blue} {\bf 819.9}^{0.9}$ \\ \hline
GPVAE\Tstrut & $\color{red} {\bf 354.8}^{1.5}$ & $\color{red} {\bf 516.3}^{2.1}$ & $\color{red} {\bf 821.8}^{4.4}$ \\
\bottomrule
\end{tabular}
\end{sc}
\end{small}
\vskip 0.05in
\end{table}
%%%%

%%%%%%%%

\textbf{Network architectures.} We adopt the convolutional neural networks for both encoder and decoder %\footnote{More precisely, we used the convolutional network for the encoder and the transposed convolution network for the decoder.} 
models for all competing approaches. The main reason is that the convolutional networks are believed to outperform fully connected networks for many tasks on the image domain\footnote{We empirically compared the two networks in the  Supplement.
%We also empirically compared the performance between the two architectures in the Supplement.
}~\cite{cnn_imagenet,cnn_szegedy,cnn_dcgan}.  %\footnote{In particular, it is shown (in the Supplement) that the fully-connected decoder architecture is inferior to the deconvolutional decoder that we adopted, when the two architectures have roughly equal numbers of parameters. This is why we exclude comparison with the recent Laplacian approximation approach of~\cite{vlae} in the main paper. They use the first-order approximation solver method to obtain the mode of the true posterior, but such linearization of a deep network is only computationally feasible for {\em fully connected} decoder models. On the other hand, our recursive mixture learning admits arbitrary types of encoder/decoder architectures, which is another advantage. In the Supplement, we report the performance of the Laplace approximation. % with fully connected decoder models.
%}. 
%
For the encoder architecture, we first apply $L$ convolutional layers with $(4 \times 4)$-pixels kernels, followed by two fully-connected layers with hidden layers dimension $h$. For the decoder, the input images first go through two fully connected layers, followed by $L$ transposed convolutional %deconvolution %\footnote{More precisely, the {\em transposed convolutional} networks.} 
layers with $(4 \times 4)$-pixels filters. Here, $L=3$ for all datasets except CelebA ($L=4$), and  $h=256$ for the MNIST/OMNIGLOT and $h=512$ for the others.
The deep kernel feature functions $\bm{\psi}^{m,s}({\bf x})$ in our GPVAE model have exactly the same architecture as the encoder network except that the last fully connected layer is removed. This ensures that the GP functions ${\bf f}({\bf x})$ and ${\bf h}({\bf x})$ have equal functional capacity to the base encoder network since they are defined to be products of the features and the Gaussian random weights ${\bf W}$ and ${\bf U}$. And, accordingly the feature dimension $p$ is set equal to $h$. The full covariance matrices of the variational density $q({\bf W},{\bf U})$ are represented by Cholesky parametrization to ensure positive definiteness (e.g., $\bm{\Sigma} = {\bf L}{\bf L}^\top$ where ${\bf L}$ is a lower triangle matrix with strictly positive diagonals). %, serving as free parameters to be learned). 

\textbf{Experimental setup}. The latent dimension is %$\textrm{dim}({\bf z})$ is %In the previous work~\cite{cremer18,vlae}, they used fully connected networks for encoder/decoder with two different complexities, small and large. But it turns out that convnet significantly outperforms the full-con nets provided that the number of parameters of the two architectures are (roughly) equal. Instead, what matters more is the latent dimension, hence we vary the $\textrm{dim}({\bf z})$ from small to large. 
chosen from $\{10,20,50\}$. To report the test log-likelihood scores, %$\log p({\bf x})$, 
we use the importance weighted %sampling
estimation (IWAE)\footnote{The details can be also found in the Supplement.}~\cite{iwae} with 100 samples. 
%
%For each model/dataset, we repeat the training procedure three times with different random seeds, and report the best results. 
For each model/dataset, we perform 10 runs with different random train/validation splits, where each run consists of three trainings by starting with different random model parameters, among which only one model with the best validation result is chosen.

%%%%
\begin{table}%[t]
\vspace{-0.6em}
\centering
\caption{(SVHN) Test log-likelihood scores (unit in nat). The same interpretation as \autoref{tab:mnist}. 
}
\label{tab:svhn}
\vspace{+0.3em}
\begin{small}
\begin{sc}
\centering
\begin{tabular}{lcccc}
\toprule
 & $\textrm{dim}({\bf z})=10$ & $\textrm{dim}({\bf z})=20$ & $\textrm{dim}({\bf z})=50$ \\
\midrule
VAE & $3360.2^{9.1}$ & $4054.5^{14.3}$ & $5363.7^{21.4}$ \\ \hline
SA$^{(1)}$\Tstrut & $3358.7^{8.9}$ & $4031.5^{19.0}$ & $5362.1^{35.7}$ \\
SA$^{(2)}$ & $3356.0^{8.8}$ & $4041.5^{15.5}$ & $5377.0^{23.2}$ \\
SA$^{(4)}$ & $3327.8^{8.2}$ & $4051.9^{22.2}$ & $5391.7^{20.4}$ \\
SA$^{(8)}$ & $3352.8^{11.5}$ & $4041.6^{9.5}$ & $5370.8^{18.5}$ \\ \hline
IAF$^{(1)}$\Tstrut & $3377.1^{8.4}$ & $4050.0^{9.4}$ & $5368.3^{11.5}$ \\
IAF$^{(2)}$ & $3362.3^{8.9}$ & $4054.6^{10.5}$ & $5360.0^{10.0}$ \\
IAF$^{(4)}$ & $3346.1^{8.7}$ & $4048.6^{8.7}$ & $5338.1^{10.2}$ \\
IAF$^{(8)}$ & $3372.6^{8.3}$ & $4042.0^{9.6}$ & $5341.8^{10.1}$ \\ \hline
HF$^{(1)}$\Tstrut & $3381.4^{8.9}$ & $4028.8^{9.7}$ & $5372.0^{10.1}$ \\
HF$^{(2)}$ & $3342.4^{8.3}$ & $4030.7^{9.9}$ & $5376.6^{10.2}$ \\
HF$^{(4)}$ & $3370.0^{8.2}$ & $4038.4^{9.7}$ & $5371.8^{9.8\ \ }$ \\
HF$^{(8)}$ & $3343.8^{8.2}$ & $4035.9^{8.9}$ & $5351.1^{11.1}$ \\ \hline
ME$^{(2)}$\Tstrut & $3352.3^{9.9}$ & $4037.2^{11.0}$ & $5343.2^{13.1}$ \\
ME$^{(3)}$ & $3335.2^{10.9}$ & $4053.8^{16.1}$ & $5367.7^{15.8}$ \\
ME$^{(4)}$ & $3358.2^{14.9}$ & $4061.3^{12.0}$ & $5191.9^{18.5}$ \\
ME$^{(5)}$ & $3360.6^{7.8}$ & $4057.5^{12.2}$ & $5209.2^{12.8}$ \\ \hline
RME$^{(2)}$\Tstrut & $3390.0^{8.1}$ & $4085.3^{9.7}$ & $\color{blue} {\bf 5403.2}^{10.2}$ \\
RME$^{(3)}$ & $3392.0^{12.6}$ & $4085.9^{9.8}$ & $\color{blue} {\bf 5405.1}^{10.4}$ \\
RME$^{(4)}$ & $3388.6^{8.3}$ & $4080.7^{9.9}$ & $\color{blue} {\bf 5403.8}^{10.2}$ \\
RME$^{(5)}$ & $3391.9^{8.2}$ & $4086.9^{10.9}$ & $\color{blue} {\bf 5405.5}^{8.5\ \ }$ \\ \hline
GPVAE\Tstrut & $\color{red} {\bf 3417.6}^{11.8}$ & $\color{red} {\bf 4133.2}^{13.3}$ & $\color{red} {\bf 5410.0}^{19.5}$ \\
\bottomrule
\end{tabular}
\end{sc}
\end{small}
\vskip 0.05in
\end{table}
%%%%

%%%%
\begin{table}[t]
\vspace{-0.6em}
\centering
\caption{(CelebA) Test log-likelihood scores (unit in nat). The same interpretation as \autoref{tab:mnist}. 
}
\label{tab:celeba}
\vspace{+0.3em}
\begin{small}
\begin{sc}
\centering
\begin{tabular}{lcccc}
\toprule
 & $\textrm{dim}({\bf z})=10$ & $\textrm{dim}({\bf z})=20$ & $\textrm{dim}({\bf z})=50$ \\
\midrule
VAE & $9767.7^{36.0}$ & $12116.4^{25.3}$ & $15251.9^{39.7}$ \\ \hline
SA$^{(1)}$\Tstrut & $9735.2^{21.4}$ & $12091.1^{21.6}$ & $15285.8^{29.4}$ \\
SA$^{(2)}$ & $9754.2^{20.4}$ & $12087.1^{21.5}$ & $15252.7^{29.0}$ \\
SA$^{(4)}$ & $9769.1^{20.6}$ & $12116.3^{20.5}$ & $15187.3^{27.9}$ \\
SA$^{(8)}$ & $9744.8^{19.4}$ & $12100.6^{22.8}$ & $15096.5^{27.2}$ \\ \hline
IAF$^{(1)}$\Tstrut & $9750.3^{27.4}$ & $12098.0^{20.6}$ & $15271.2^{28.6}$ \\
IAF$^{(2)}$ & $9794.4^{23.3}$ & $12104.5^{21.8}$ & $15262.2^{27.8}$ \\
IAF$^{(4)}$ & $9764.7^{29.5}$ & $12094.6^{22.6}$ & $15261.0^{28.1}$ \\
IAF$^{(8)}$ & $9764.0^{21.6}$ & $12109.3^{22.0}$ & $15241.5^{27.9}$ \\ \hline
HF$^{(1)}$\Tstrut & $9748.3^{29.5}$ & $12077.2^{31.4}$ & $15240.5^{27.6}$ \\
HF$^{(2)}$ & $9765.8^{25.6}$ & $12093.0^{25.6}$ & $15258.2^{30.3}$ \\
HF$^{(4)}$ & $9754.3^{23.8}$ & $12082.0^{27.0}$ & $15266.5^{29.5}$ \\
HF$^{(8)}$ & $9737.5^{24.5}$ & $12087.3^{25.5}$ & $15248.7^{29.7}$ \\ \hline
ME$^{(2)}$\Tstrut & $\color{blue} {\bf 9825.3}^{20.7}$ & $12072.7^{23.3}$ & $15290.5^{29.3}$ \\
ME$^{(3)}$ & $9797.6^{22.3}$ & $12100.3^{21.7}$ & $15294.6^{28.3}$ \\
ME$^{(4)}$ & $\color{blue} {\bf 9834.9}^{25.4}$ & $12092.2^{22.6}$ & $15270.7^{20.6}$ \\
ME$^{(5)}$ & $9717.0^{23.2}$ & $12095.3^{25.1}$ & $15268.8^{27.5}$ \\ \hline
RME$^{(2)}$\Tstrut & $\color{blue} {\bf 9837.9}^{24.6}$ & $\color{blue} {\bf 12193.1}^{23.5}$ & $\color{blue} {\bf 15363.0}^{31.7}$ \\
RME$^{(3)}$ & $\color{blue} {\bf 9838.5}^{25.0}$ & $\color{blue} {\bf 12192.3}^{23.5}$ & $\color{blue} {\bf 15365.6}^{31.4}$ \\
RME$^{(4)}$ & $\color{red} {\bf 9849.5}^{12.1}$ & $\color{blue} {\bf 12192.6}^{23.4}$ & $\color{blue} {\bf 15364.3}^{31.5}$ \\
RME$^{(5)}$ & $\color{blue} {\bf 9843.5}^{25.0}$ & $\color{red} {\bf 12194.2}^{11.5}$ & $\color{blue} {\bf 15366.2}^{12.7}$ \\ \hline
GPVAE\Tstrut & $\color{blue} {\bf 9843.4}^{23.8}$ & $\color{blue} {\bf 12184.0}^{22.7}$ & $\color{red} {\bf 15370.9}^{27.6}$ \\
\bottomrule
\end{tabular}
\end{sc}
\end{small}
\vskip 0.05in
\end{table}
%%%%

%%%%%%%%%%%%%%%%%%%%%%%%%%%%%%%%%%%%%%%%%%%%%%%%%%%%%%%%%%%%%%%%%%%%%%%%%%%%%%%
\subsection{Results}\label{sec:results}

The test log-likelihood scores are summarized in \autoref{tab:mnist} (MNIST), \autoref{tab:omniglot} (OMNIGLOT), %\autoref{tab:cifar10} (CIFAR10), 
\autoref{tab:svhn} (SVHN), and \autoref{tab:celeba} (CelebA). %The qualitative results of reconstructed and synthesized images can be found in the Supplement. 
% Analysis/interpretation for each dataset...
Our GPVAE overall outperforms the competing approaches consistently for all datasets. Below we provide interpretation for the results.

%\textbf{Comparison to SA.} 
The performance of the semi-amortized approach (SA) is mixed, sometimes achieving improvement over VAE, but not consistently. SA's performance is very sensitive to the number of SVI gradient update steps, another drawback of the SA where the gradient-based adaption has to be performed at test time. Although one could adjust the gradient step size (currently we are using a fixed gradient step size) to improve the performance, as far as we know, there is little  principled way to tune the step size at test time that can attain optimal accuracy and inference time trade off. 

%\textbf{Comparison to IAF/HF.} 
The flow-based models (IAF and HF) adopt nonlinear invertible transformations to enrich the representational capacity of the variational posterior. % over the VAE's post-Gaussian. 
In principle, they are capable of representing highly nonlinear non-Gaussian conditional densities, perhaps subsuming the true posteriors, via autoregressive flows (IAF) and the Householder transformed full covariance matrices (HF). However, their improvement in accuracy over the VAE trails that of our GPVAE; they often perform only as well as the VAE. %with the tendency to possibly overfit. 
The failure of the flow-based models might be due to the difficulty in optimizing complex encoder models where similar observations were made in related previous work~\cite{vlae,mkim2020neurips}. This result suggests that sophisticated and discriminative learning criteria are critical, beyond just enlarging the structural capacity of the neural networks. Our GPVAE's explicit modeling of the deviation of the base encoder density from the true posterior via GP noise processes accomplishes this goal.

%\textbf{Comparison to ME.} 
Similarly, despite its increased functional capacity, the mixture encoder (ME) also has difficulty in learning a good model, being quite sensitive to the initial parameters. Except for one case on CelebA with $\textrm{dim}({\bf z})=10$, it consistently underperforms our GPVAE. The blind mixture estimation can potentially suffer from collapsed mixture components and dominant single component issues. The fact that even the baseline VAE often performs comparably to the ME with different mixture orders supports this observation. This, again, signifies the importance of employing more discriminative learning criteria, as done by our GPVAE's explicit modeling of the posterior deviation. The RME, by following this direction of adopting a discriminative learning objective, performs equally well with our GPVAE on many cases, but slightly underperforms ours on the others.

%%%%
\begin{figure*}%[t!]
%\vspace{-1.0em}
\begin{center}
\includegraphics[trim = 0mm 0mm 0mm 0mm, clip, scale=0.415%0.445%0.425
]{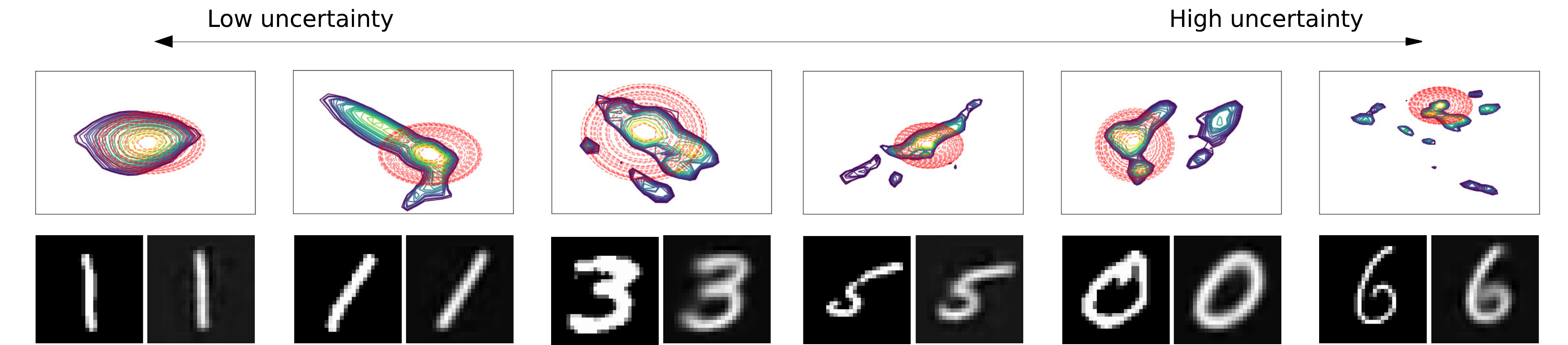}
\end{center}
\vspace{-1.5em}
\caption{
Uncertainty vs.~posterior approximation difficulty. After the GPVAE model is trained on MNIST with 2D latent space, we evaluate the  uncertainty $\textrm{Tr}\mathbb{V}({\bf f}({\bf x})|\mathcal{D}) + \textrm{Tr}\mathbb{V}({\bf h}({\bf x})|\mathcal{D})$, and depict six different instances ${\bf x}$ in the order of increasing uncertainty values. Top panel shows the true posterior $p_{\bm{\theta}}({\bf z}|{\bf x})$ %\hl{you mean $\propto p_{\bm{\theta}}({\bf x}|\bm{z})$?} 
(contour plots) and the base encoder $\mathcal{N}({\bf b}({\bf x}),{\bf c}({\bf x})^2)$ (red dots) superimposed (in log scale). Bottom panel contains the original inputs ${\bf x}$ (left) and reconstructed images (right). For the cases with lower uncertainty, the true posteriors are more Gaussian-like. %, and the reconstruction quality is high. 
On the other hand, the higher uncertainty cases have highly non-Gaussian true posteriors with multiple modes. %; the subjective reconstruction quality degrades more severely. 
}
\label{fig:uncertainty_vs_difficulty}
%\vspace{-0.5em}
\vskip 0.05in
\end{figure*}
%%%%

%%%%%%%%%%%%%%%%%%%%%%%%%%%%%%%%%%%%%%%%%%%%%%%%%%%%%%%%%%%%%%%%%%%%%%%%%%%%%%%
\subsection{Test Inference Time}\label{sec:inf_time}

Compared to the semi-amortized methods, the inference in our GP encoder network is much faster as it is accomplished by a single feed forward pass through the encoder network. Unlike the semi-amortized approaches where one has to perform the SVI gradient adaptation at test time, in our GPVAE model, after the training stage, the posterior model is fixed, with no further adaptation required. 
%
%%%%
\begin{table}%[t]
\vspace{-0.6em}
\centering
\caption{(Per-batch) Test inference time (unit in milliseconds) with batch size 128. The latent dimension $\textrm{dim}({\bf z})=50$. %MNIST & OMNIGLOT & CIFAR10 & SVHN & CelebA
}
\label{tab:inf_time}
\vspace{+0.3em}
\begin{small}
\begin{sc}
\centering
\begin{tabular}{lccccc}
\toprule
%& M & O & Ci & S & Ce \\
& \scriptsize{MNIST} & \scriptsize{OMNIG.} & \scriptsize{CIFAR10} & \scriptsize{SVHN} & \scriptsize{CelebA} \\
\midrule
VAE & \ \ 3.6 & \ \ 4.8 & \ \ 3.7 & \ \ 2.2 & \ \ 2.7 \\ \hline
SA$^{(1)}$\Tstrut & \ \ 9.7 & 11.6 & \ \ 9.8 & \ \ 7.0 & \ \ 8.4 \\
SA$^{(2)}$ & 18.1 & 19.2 & 16.8 & 15.5 & 13.8 \\
SA$^{(4)}$ & 32.2 & 34.4 & 27.9 & 30.1 & 27.1 \\
SA$^{(8)}$ & 60.8 & 65.7 & 60.5 & 60.3 & 53.8 \\ \hline
IAF$^{(1)}$\Tstrut & \ \ 4.8 & \ \ 5.7 & \ \ 5.1 & \ \ 3.4 & \ \ 4.4 \\
IAF$^{(2)}$ & \ \ 5.9 & \ \ 6.4 & \ \ 5.6 & \ \ 3.7 & \ \ 5.1 \\
IAF$^{(4)}$ & \ \ 6.2 & \ \ 7.0 & \ \ 6.3 & \ \ 4.7 & \ \ 5.7 \\
IAF$^{(8)}$ & \ \ 7.7 & \ \ 8.2 & \ \ 7.6 & \ \ 5.7 & \ \ 7.7 \\ \hline
GPVAE\Tstrut & \ \ 9.9 & 10.2 & \ \ 9.3 & \ \ 8.0 & \ \ 9.2 \\
\bottomrule
\end{tabular}
\end{sc}
\end{small}
\vskip 0.05in
\end{table}
%%%%
%
To verify computational speed-up over the semi-amortized approaches and others, we measure the inference. The per-batch inference times (batch size 128) on all benchmark datasets are shown in \autoref{tab:inf_time}. To report the results, for each method and each dataset, we run inference over the entire test set batches, measure the running time, then take the per-batch average. We repeat the procedure five times and report the average. All models are run on the same machine with a single GPU (RTX 2080 Ti), Core i7 3.50GHz CPU, and 128 GB RAM. 

Note that we only report test times for the latent dimension $\textrm{dim}({\bf z})=50$ since the impact of the latent dimension appears to be less significant for all models except for our GPVAE. 
In the GPVAE, the latent dimension can considerably affect the inference time because of the matrix operations performed per latent dimension (c.f., (\ref{eq:qzx_mm})). Hence, we consider the most complex (worst) case for our GPVAE model, $\textrm{dim}({\bf z})=50$, the highest dimension in our experimental setup. 
Most notably yet as expected, the semi-amortized approach (SA) suffers from the computational overhead of test time gradient updates, with the inference time significantly growing as a function of the number of increasing updates. Our GPVAE is significantly faster than the SA with more than one SVI step, albeit on par or slower than the flow-based IAF. We believe that the inference time of the GPVAE can be further improved by more effective implementations of the dimension-wise matrix operations, which remain as our future work\footnote{Our current implementation uses \texttt{for} loop to iterate matrix operations over latent dimensions, but can be potentially converted to block operations without a loop, possibly with parallelization.
}.

%%%%%%%%%%%%%%%%%%%%%%%%%%%%%%%%%%%%%%%%%%%%%%%%%%%%%%%%%%%%%%%%%%%%%%%%%%%%%%%
\subsection{Uncertainty vs.~Posterior Approximation Difficulty}\label{sec:uncertainty}

Another important benefit of our Bayesian treatment is that we can quantify the {\em uncertainty} %difficulty (or discrepancy) 
in posterior approximation. Recall that our GP posterior $p({\bf f},{\bf h}|\mathcal{D})$ captures the discrepancy between the base encoder $\mathcal{N}({\bf b}({\bf x}),{\bf c}({\bf x})^2)$ and the true posterior $p_{\bm{\theta}}({\bf x})$ via the GP noise processes ${\bf f}({\bf x})$ and ${\bf h}({\bf x})$. In particular, the variance $\mathbb{V}(f({\bf x})|\mathcal{D})$) (similarly for ${\bf h}$) at given input ${\bf x}$, can serve as a useful indicator that gauges the goodness of posterior approximation via a single Gaussian. For instance,  the large posterior variance ({\em high uncertainty}) implies that the posterior approximation is {\em difficult}, suggesting the true posterior is distinct from a Gaussian (e.g., having multiple modes). On the other hand, if the variance is small ({\em low uncertainty}), one can anticipate that the true posterior might be close to a Gaussian. \autoref{fig:uncertainty_vs_difficulty} illustrates this intuition on the MNIST dataset with 2D latent space, where the uncertainty measured by $\textrm{Tr}\mathbb{V}({\bf f}({\bf x})|\mathcal{D}) + \textrm{Tr}\mathbb{V}({\bf h}({\bf x})|\mathcal{D})$ accurately aligns with the non-Gaussianity of the true posterior, closely related to the quality of reconstruction. % (when visually inspected). 

%%%%%%%%%%%%%%%%%%%%%%%%%%%%%%%%%%%%%%%%%%%%%%%%%%%%%%%%%%%%%%%%%%%%%%%%%%%%%%%
%%%%%%%%%%%%%%%%%%%%%%%%%%%%%%%%%%%%%%%%%%%%%%%%%%%%%%%%%%%%%%%%%%%%%%%%%%%%%%%
\section{Conclusions}\label{sec:conclusion}

%In this paper 
We have proposed a novel Gaussian process encoder model to significantly reduce the posterior approximation error of the amortized inference in VAE, while being computationally more efficient than recent semi-amortized approaches. Our Bayesian treatment that regards the discrepancy in posterior approximation as a random noise process, leads to improvements in the accuracy of inference within the fast amortized inference framework. It also offers the ability to quantify the uncertainty in variational inference, intuitively interpreted as inherent difficulty in posterior approximation. In our future work, we plan to apply GPVAE to domains with structured data, including sequences %(e.g., videos and natural language sentences) 
and graphs. % (e.g., molecules). % and 3D shape data), which can be effectively represented within the GP framework. 

%%%%%%%%%%%%%%%%%%%%%%%%%%%%%%%%%%%%%%%%%%%%%%%%%%%%%%%%%%%%%%%%%%%%%%%%%%%%%%%
%%%%%%%%%%%%%%%%%%%%%%%%%%%%%%%%%%%%%%%%%%%%%%%%%%%%%%%%%%%%%%%%%%%%%%%%%%%%%%%
%\section*{References}

%\small
{
\bibliography{main}
\bibliographystyle{icml2021}
}

%%%%%%%%%%%%%%%
%\subsubsection*{Acknowledgments}

%Use unnumbered third level headings for the acknowledgments. 

\newpage 

\begin{center}
%\vspace*{\fill}
\LARGE Supplementary Material
%\vspace*{\fill}
\end{center}

%%%%%%%%%%%%%%%%%%%%%%%%%%%%%%%%%%%%%%%%%%%%%%%%%%%%%%%%%%%%%%%%%%%%%%%%%%%%%%%
%%%%%%%%%%%%%%%%%%%%%%%%%%%%%%%%%%%%%%%%%%%%%%%%%%%%%%%%%%%%%%%%%%%%%%%%%%%%%%%
\section{Detailed Derivations for GP Inference and Learning}\label{sec:derivations}

We provide detailed derivations for the variational inference for GP (Sec.~3.3 in the main paper). Specifically, we show that
%%%%
\begin{equation}
\textrm{KL} \big( q({\bf W},{\bf U}) || p({\bf W},{\bf U}|\mathcal{D}) \big) = \log \hat{p}_{\bm{\theta}}(\mathcal{D}) - \sum_{{\bf x} \in\mathcal{D}} \textrm{ELBO}(\bm{\theta}, \bm{\Lambda}; {\bf x}),
\label{eq:kl_elbo}
\end{equation}
%%%%
where $\hat{p}_{\bm{\theta}}(\mathcal{D}) = \mathbb{E}_{{\bf W},{\bf U} \sim \mathcal{N}({\bf 0}, {\bf I})} \big[ \prod_{{\bf x}} e^{\mathcal{L}_{\bm{\theta}}({\bf W}, {\bf U}; {\bf x})} \big]$ is the marginal data likelihood using the surrogate likelihood function $\mathcal{L}_{\bm{\theta}}({\bf W}, {\bf U}; {\bf x}) := \mathbb{E}_{q} \Big[ \log \frac{ p_{\bm{\theta}}({\bf x},{\bf z}) }{ q({\bf z}|{\bf x},{\bf W},{\bf U}) } \Big]$, and 
%%%%
\begin{equation}
\textrm{ELBO}(\bm{\theta}, \bm{\Lambda}; {\bf x}) = 
\mathbb{E}_{q({\bf z}|{\bf x})} \big[ \log p_{\bm{\theta}}({\bf x}|{\bf z}) \big] - \mathbb{E}_{q({\bf W},{\bf U})}[ \textrm{KL}( q({\bf z}|{\bf x},{\bf W},{\bf U}) || p({\bf z}) ) ] - \frac{1}{N} \textrm{KL}( q({\bf W},{\bf U}) || \mathcal{N}({\bf 0}, {\bf I}) ).
\end{equation}
%%%%

\textbf{Proof.} Starting from the left hand side of (\ref{eq:kl_elbo}),
%%%%
\begin{align}
&\textrm{KL} \big( q({\bf W},{\bf U}) || p({\bf W},{\bf U}|\mathcal{D}) \big) \ = \
\mathbb{E}_{q({\bf W},{\bf U})} \Big[ 
   \log \frac{q({\bf W},{\bf U})}{p({\bf W},{\bf U}|\mathcal{D})}
\Big] \\
& \ \ = \
\mathbb{E}_{q({\bf W},{\bf U})} \Bigg[ 
   \log \frac{ q({\bf W},{\bf U}) \    
       \hat{p}_{\bm{\theta}}(\mathcal{D}) }
   { \mathcal{N}({\bf W},{\bf U};{\bf 0}, {\bf I}) 
     \prod_{{\bf x}\in\mathcal{D}} e^{\mathcal{L}_{\bm{\theta}}({\bf W}, {\bf U}; {\bf x})}
   }
\Bigg] \\
& \ \ = \
\log \hat{p}_{\bm{\theta}}(\mathcal{D}) + 
  \textrm{KL}( q({\bf W},{\bf U}) || \mathcal{N}({\bf 0}, {\bf I}) ) - 
  \sum_{{\bf x}\in\mathcal{D}} 
      \mathbb{E}_{q({\bf W},{\bf U})} \big[ 
        \mathcal{L}_{\bm{\theta}}({\bf W}, {\bf U}; {\bf x})
      \big] \\
& \ \ = \
\log \hat{p}_{\bm{\theta}}(\mathcal{D}) + 
  \textrm{KL}( q({\bf W},{\bf U}) || \mathcal{N}({\bf 0}, {\bf I}) ) - 
  \sum_{{\bf x}\in\mathcal{D}}  \bigg(
      \mathbb{E}_{q({\bf z}|{\bf x})} \big[ \log p_{\bm{\theta}}({\bf x}|{\bf z}) \big] - 
      \mathbb{E}_{q({\bf W},{\bf U})} \big[ 
        \textrm{KL}( q({\bf z}|{\bf x},{\bf W},{\bf U}) || p({\bf z}) )
      \big]
  \bigg).
\end{align}
%%%%
For the last equality we use the GP posterior marginalized encoder distribution, 
%%%%
\begin{equation}
q({\bf z}|{\bf x}) = \iint q({\bf W},{\bf U}) q({\bf z}|{\bf x},{\bf W},{\bf U}) d {\bf W} d{\bf U}.
\label{eq:qzx}
\end{equation}
%%%%
Arranging the last equation completes the proof.

%%%%%%%%%%%%%%%%%%%%%%%%%%%%%%%%%%%%%%%%%%%%%%%%%%%%%%%%%%%%%%%%%%%%%%%%%%%%%%%
%%%%%%%%%%%%%%%%%%%%%%%%%%%%%%%%%%%%%%%%%%%%%%%%%%%%%%%%%%%%%%%%%%%%%%%%%%%%%%%
\section{Experimental Setups and Network Architectures}\label{sec:expmt_setup}

For all optimization, we used the Adam optimizer with batch size $128$ and learning rate $0.0005$. We run the optimization until 2000 epochs. We use the same encoder/decoder architectures for competing methods including our GPVAE, VAE, SA (semi-amortized approach), and also the base density in the flow-based models IAF and HF. The network architectures are slightly different across the datasets due to different input image dimensions. We summarize the full network architectures in \autoref{tab:arch_1x28x28} (MNIST and OMNIGLOT), \autoref{tab:arch_3x32x32} (CIFAR10 and SVHN), and \autoref{tab:arch_3x64x64} (CelebA).

%%%%%%%%
\begin{table*}[h!]
\caption{Encoder and decoder network architectures for MNIST and OMNIGLOT datasets. In the convolutional and transposed convolutional layers, the paddings are properly adjusted to match the input/output dimensions.}
\label{tab:arch_1x28x28}
\vskip 0.05in
\begin{center}
\begin{small}
\begin{sc}
\begin{tabular}{l|l}
\toprule
 Encoder & Decoder \\
\midrule
Input: $(28 \times 28 \times 1)$ & Input: ${\bf z} \in $ $\mathbb{R}^{p}$ ($p\in\{10,20,50\})$ \\
\midrule
32 (4 $\times$ 4) conv.; stride 2; LeakyReLU ($0.01$) & FC. 256; ReLU \\
\midrule
32 (4 $\times$ 4) conv.; stride 2; LeakyReLU ($0.01$) & FC. $3 \cdot 3 \cdot 64$; RELU \\
\midrule
64 (4 $\times$ 4) conv.; stride 2; LeakyReLU ($0.01$) & 32 (4 $\times$ 4) Transposed Conv.; stride 2; ReLU \\
\midrule
FC. 256; LeakyReLU ($0.01$) & 32 (4 $\times$ 4) Transposed Conv.; stride 2; ReLU \\
\midrule
FC. 2 $\times p$ ($p=\textrm{dim}({\bf z})\in\{10,20,50\})$ & 1 (4 $\times$ 4) Transposed Conv.; stride 2 \\
\bottomrule
\end{tabular}
\end{sc}
\end{small}
\end{center}
\vskip -0.1in
%\vspace{-1.0em}
\end{table*}
%%%%%%%%

%%%%%%%%
\begin{table*}[h!]
\caption{Encoder %(i.e., each component in our mixture model)
and decoder network architectures for CIFAR10 and SVHN datasets. %In the convolutional and transposed convolutional layers, the paddings are properly adjusted to match the input/output dimensions.
}
\label{tab:arch_3x32x32}
\vskip 0.05in
\begin{center}
\begin{small}
\begin{sc}
\begin{tabular}{l|l}
\toprule
 Encoder & Decoder \\
\midrule
Input: $(32 \times 32 \times 3)$ & Input: ${\bf z} \in $ $\mathbb{R}^{p}$ ($p\in\{10,20,50\})$ \\
\midrule
32 (4 $\times$ 4) conv.; stride 2; LeakyReLU ($0.01$) & 
FC. 512; ReLU \\
\midrule
32 (4 $\times$ 4) conv.; stride 2; LeakyReLU ($0.01$) & 
FC. $4 \cdot 4 \cdot 64$; RELU \\
\midrule
64 (4 $\times$ 4) conv.; stride 2; LeakyReLU ($0.01$) & 
32 (4 $\times$ 4) Transposed Conv.; stride 2; ReLU \\
\midrule
FC. 512; LeakyReLU ($0.01$) & 
32 (4 $\times$ 4) Transposed Conv.; stride 2; ReLU \\
\midrule
FC. 2 $\times p$ ($p=\textrm{dim}({\bf z})\in\{10,20,50\})$ & 
3 (4 $\times$ 4) Transposed Conv.; stride 2 \\
\bottomrule
\end{tabular}
\end{sc}
\end{small}
\end{center}
\vskip -0.1in
%\vspace{-1.0em}
\end{table*}
%%%%%%%%

%%%%%%%%
\begin{table*}[h!]
\caption{Encoder %(i.e., each component in our mixture model)
and decoder network architectures for CelebA dataset. %In the convolutional and transposed convolutional layers, the paddings are properly adjusted to match the input/output dimensions.
}
\label{tab:arch_3x64x64}
\vskip 0.05in
\begin{center}
\begin{small}
\begin{sc}
\begin{tabular}{l|l}
\toprule
 Encoder & Decoder \\
\midrule
Input: $(64 \times 64 \times 3)$ & Input: ${\bf z} \in $ $\mathbb{R}^{p}$ ($p\in\{10,20,50\})$ \\
\midrule
32 (4 $\times$ 4) conv.; stride 2; LeakyReLU ($0.01$) & 
FC. 512; ReLU \\
\midrule
32 (4 $\times$ 4) conv.; stride 2; LeakyReLU ($0.01$) & 
FC. $4 \cdot 4 \cdot 64$; RELU \\
\midrule
64 (4 $\times$ 4) conv.; stride 2; LeakyReLU ($0.01$) & 
64 (4 $\times$ 4) Transposed Conv.; stride 2; ReLU \\
\midrule
64 (4 $\times$ 4) conv.; stride 2; LeakyReLU ($0.01$) & 
32 (4 $\times$ 4) Transposed Conv.; stride 2; ReLU \\
\midrule
FC. 512; LeakyReLU ($0.01$) & 
32 (4 $\times$ 4) Transposed Conv.; stride 2; ReLU \\
\midrule
FC. 2 $\times p$ ($p=\textrm{dim}({\bf z})\in\{10,20,50\})$ & 
3 (4 $\times$ 4) Transposed Conv.; stride 2 \\
\bottomrule
\end{tabular}
\end{sc}
\end{small}
\end{center}
%\vskip -0.1in
%\vspace{-1.0em}
\end{table*}
%%%%%%%%

%%%%%%%%%%%%%%%%%%%%%%%%%%%%%%%%%%%%%%%%%%%%%%%%%%%%%%%%%%%%%%%%%%%%%%%%%%%%%%%
%%%%%%%%%%%%%%%%%%%%%%%%%%%%%%%%%%%%%%%%%%%%%%%%%%%%%%%%%%%%%%%%%%%%%%%%%%%%%%%
\section{Additional Experiments on CIFAR10}\label{sec:extra_cifar10}

The results on the CIFAR10 dataset are reported in Tab~\ref{tab:cifar10}. 

\definecolor{lor}{rgb}{1,0.85,0}
\definecolor{or}{rgb}{1,0.60,0}
\definecolor{dor}{rgb}{1,0.20,0}
% \newcommand\Tstrut{\rule{0pt}{2.2ex}}         % = `top' strut
% \newcommand\Bstrut{\rule[-0.9ex]{0pt}{0pt}}   % = `bottom' strut

%%%%
\begin{table}[t]
\vspace{-0.6em}
\centering
\caption{(CIFAR10) Test log-likelihood scores (unit in nat) estimated by the importance weighted sampling~\cite{iwae} with 100 samples. 
The figures in the parentheses next to model names indicate: the number of SVI steps in SA,  the number of flows in IAF and HF, and the number of mixture components in ME and RME. The superscripts are the standard deviations. 
The best (on average) results are boldfaced in $\color{red} \textrm{\textbf{red}}$. In each column, we perform the two-sided $t$-test to measure the statistical significance of the difference between the best model (red) and each competing method. We depict those with $p$-values greater than $0.01$ as boldfaced $\color{blue} \textrm{blue}$ (little evidence of difference). So, anything plain non-colored indicates $p\leq 0.01$ (significantly different). 
Best viewed in color.
}
\label{tab:cifar10}
\vspace{+0.3em}
\begin{small}
\begin{sc}
\centering
\begin{tabular}{lcccc}
\toprule
 & $\textrm{dim}({\bf z})=10$ & $\textrm{dim}({\bf z})=20$ & $\textrm{dim}({\bf z})=50$ \\
\midrule
VAE & $1645.7^{4.9}$ & $2089.7^{5.8}$ & $2769.9^{7.1}$ \\ \hline
SA$^{(1)}$\Tstrut & $1645.0^{5.6}$ & $2086.0^{6.2}$ & $2765.0^{7.1}$ \\
SA$^{(2)}$ & $1648.6^{4.8}$ & $2088.2^{6.6}$ & $2764.1^{7.7}$ \\
SA$^{(4)}$ & $1648.5^{5.2}$ & $2083.9^{8.4}$ & $2766.7^{6.6}$ \\
SA$^{(8)}$ & $1642.1^{5.4}$ & $2086.0^{6.1}$ & $2766.6^{7.5}$ \\ \hline
IAF$^{(1)}$\Tstrut & $1646.0^{4.9}$ & $2081.1^{5.4}$ & $2762.6^{7.2}$ \\
IAF$^{(2)}$ & $1642.0^{4.9}$ & $2084.6^{5.6}$ & $2763.0^{4.3}$ \\
IAF$^{(4)}$ & $1646.0^{5.1}$ & $2083.2^{6.1}$ & $2760.6^{7.0}$ \\
IAF$^{(8)}$ & $1643.6^{4.6}$ & $2087.1^{4.6}$ & $2761.8^{6.9}$ \\ \hline
HF$^{(1)}$\Tstrut & $1644.5^{4.4}$ & $2079.1^{5.5}$ & $2757.9^{4.4}$ \\
HF$^{(2)}$ & $1636.7^{4.9}$ & $2086.0^{5.9}$ & $2764.7^{4.4}$ \\
HF$^{(4)}$ & $1642.1^{4.9}$ & $2082.3^{7.3}$ & $2763.4^{4.4}$ \\
HF$^{(8)}$ & $1639.9^{5.4}$ & $2084.7^{6.1}$ & $2765.5^{7.2}$ \\ \hline
ME$^{(2)}$\Tstrut & $1643.6^{5.1}$ & $2086.6^{6.8}$ & $2767.9^{9.4}$ \\
ME$^{(3)}$ & $1638.6^{5.8}$ & $2079.8^{5.9}$ & $2770.2^{7.8}$ \\
ME$^{(4)}$ & $1641.8^{5.4}$ & $2084.7^{6.9}$ & $2763.5^{9.3}$ \\
ME$^{(5)}$ & $1641.7^{5.6}$ & $2080.2^{5.9}$ & $2766.1^{6.3}$ \\ \hline
RME$^{(2)}$\Tstrut & $1652.3^{5.0}$ & $2095.7^{5.8}$ & $2779.6^{6.6}$ \\
RME$^{(3)}$ & $1654.2^{4.9}$ & $2099.1^{7.2}$ & $\color{blue} {\bf 2783.0}^{6.1}$ \\
RME$^{(4)}$ & $1655.0^{6.4}$ & $2096.6^{5.9}$ & $\color{blue} {\bf 2781.1}^{6.6}$ \\
RME$^{(5)}$ & $1654.5^{4.6}$ & $2098.4^{5.8}$ & $\color{blue} {\bf 2782.9}^{6.4}$ \\ \hline
GPVAE\Tstrut & $\color{red} {\bf 1668.6}^{5.6}$ & $\color{red} {\bf 2113.0}^{6.7}$ & $\color{red} {\bf 2786.5}^{8.2}$ \\
\bottomrule
\end{tabular}
\end{sc}
\end{small}
\end{table}
%%%%

%%%%%%%%%%%%%%%%%%%%%%%%%%%%%%%%%%%%%%%%%%%%%%%%%%%%%%%%%%%%%%%%%%%%%%%%%%%%%%%
%%%%%%%%%%%%%%%%%%%%%%%%%%%%%%%%%%%%%%%%%%%%%%%%%%%%%%%%%%%%%%%%%%%%%%%%%%%%%%%
\section{Importance Weighted Sampling
Estimation (IWAE)}\label{sec:iwae}

To report the test log-likelihood scores, $\log p({\bf x})$, we used the importance weighted sampling estimation (IWAE)~\cite{iwae}. More specifically, 
%%%%
%\begin{equation}
$\textrm{IWAE} = \log %\Bigg( 
\frac{1}{K} \sum_{i=1}^K 
  \frac{ p({\bf x},{\bf z}_i) } { q({\bf z}_i|{\bf x}) } 
%\Bigg)
  $,
%\label{eq:iwae}
%\end{equation}
%%%%
where ${\bf z}_1,\dots,{\bf z}_K$ are i.i.d.~samples from $q({\bf z}|{\bf x})$. It can be shown that IWAE lower bounds $\log p({\bf x})$ and can be arbitrarily close to the target as the number of samples $K$ goes large. We used $K=100$ throughout the experiments.

%%%%%%%%%%%%%%%%%%%%%%%%%%%%%%%%%%%%%%%%%%%%%%%%%%%%%%%%%%%%%%%%%%%%%%%%%%%%%%%
%%%%%%%%%%%%%%%%%%%%%%%%%%%%%%%%%%%%%%%%%%%%%%%%%%%%%%%%%%%%%%%%%%%%%%%%%%%%%%%
\section{Fully-Connected Decoder Networks}\label{sec:fullcon_decoder}

% \hl{In particular,} it is shown (in the Supplement) that the fully-connected decoder architecture is inferior to the deconvolutional decoder that we adopted, when the two architectures have roughly equal numbers of parameters. This is why we exclude comparison with the recent Laplacian approximation approach of~\cite{vlae} in the main paper. They use the first-order approximation solver method to obtain the mode of the true posterior, but such linearization of a deep network is only computationally feasible for {\em fully connected} decoder models. On the other hand, our recursive mixture learning admits arbitrary types of encoder/decoder architectures, which is another advantage. In the Supplement, we report the performance of the Laplace approximation. % with fully connected decoder models.

In the main paper we used the convolutional networks for both encoder and decoder models. This is a reasonable architectural choice considering that all the datasets are images. It is also widely believed that convolutional networks outperform fully connected networks for many tasks in the image domain~\cite{cnn_imagenet,cnn_szegedy,cnn_dcgan}. One can alternatively consider fully connected networks for either the encoder or the decoder, or both. Nevertheless, with the equal number of model parameters, having both convolutional encoder and decoder networks always outperformed the fully connected counterparts. %More specifically, the model equipped with fully connected encoder and decoder networks is the worst, then the model having convolutional network for either encoder of decoder is better, and using convnets for both encoder and decoder performs the best.  
In this section we empirically verify this by comparing the test likelihood performance between the two architectures. We particularly focus on comparing two architectures (convolutional vs.~fully connected) for the {\em decoder} model alone, while retaining the convolutional network structure of encoders for both cases. 

Using the fully connected decoder network allows us to test the recent Laplacian approximation approach~\cite{vlae} (denoted by \textbf{VLAE}), which we excluded from the main paper. They use a first-order approximation solver to find the mode of the true posterior (i.e., linearizing the decoder function), and compute the Hessian of the log-posterior at the mode to define the  (full) covariance matrix. This procedure is  computationally feasible only for a fully connected decoder model. We conduct experiments on MNIST and OMNIGLOT datasets where the fully connected decoder network consists of two hidden layers and the hidden layer dimensions are chosen in a manner where the total number of weight parameters is roughly equal to the convolutional decoder network that we used in the main paper. 

\autoref{tab:fullcon_decoder} summarizes the results. Among the fully connected networks, the VLAE achieves the highest performance. Instead of doing SVI gradient updates as in the SAVI method (SA), the VLAE aims to directly solve the mode of the true posterior by decoder linearization, leading to more accurate posterior refinement without suffering from the step size issue. % (and the Hessian of the true negative log posterior at the mode point), and set them as the mean and covariance of the encoder. The mode seeking is done by the first-order approximate solver method, which somehow circumvents the SAVI's step size issue. 
Our GPVAE, with the fully connected decoder networks still improves the VAE's scores, but the improvement is often less than that of the VLAE. However, when compared to the convnet decoder cases, even the conventional VAE significantly outperforms the VLAE in the ability to represent the data (markedly higher log-likelihood scores). The best VLAE's scores are lower than even VAE's using convolutional decoder network. Restricted network architecture in the VLAE is its main drawback. 

%%%%
\begin{table}%[t]
\centering
\caption{%(OMNIGLOT with 
(Fully connected vs.~convolutional decoder networks) Test log-likelihood scores (unit in nat). The figures without parentheses are the scores using the fully connected networks, whereas figures in the parentheses are the scores using the convolutional decoder networks. Both architectures have roughly equal number of weight parameters. The number of linearization steps in the VLAE is chosen among $\{1,2,4,8\}$.
}
\label{tab:fullcon_decoder}
\vskip 0.05in
\begin{small}
\begin{sc}
\centering
\begin{tabular}{l|cc|cc}
\toprule
 & \multicolumn{2}{c}{MNIST} & \multicolumn{2}{|c}{OMNIGLOT} \\
 \cline{2-5}
 & \ \ $\textrm{dim}({\bf z})=10$ \ \ & \ \ $\textrm{dim}({\bf z})=50$ \ \ & \ \ $\textrm{dim}({\bf z})=10$ \ \ & \ \ $\textrm{dim}({\bf z})=50$ \ \ \\
\midrule
VAE & 563.6 (685.1) & 872.6 (1185.7) & 296.8 (347.0) & 519.4 (801.6) \\ \hline
SA ($1$) & 565.1 (688.1) & 865.8 (1172.1) & 297.6 (344.1) & 489.0 (792.7) \\
SA ($2$) & 565.3 (682.2) & 868.2 (1176.3) & 295.3 (349.5) & 534.1 (793.1) \\
SA ($4$) & 565.9 (683.5) & 852.9 (1171.3) & 294.8 (342.1) & 497.8 (794.4) \\
SA ($8$) & 564.9 (684.6) & 870.9 (1183.2) & 299.0 (344.8) & 500.0 (799.4) \\ \hline
VLAE ($1$) & 590.0 & 922.2 & 307.4 & 644.0 \\
VLAE ($2$) & 595.1 & 908.8 & 307.6 & 621.4 \\
VLAE ($4$) & 605.2 & 841.4 & 318.0 & 597.7 \\
VLAE ($8$) & 605.7 & 779.9 & 316.6 & 553.1  \\ \hline
GPVAE & 568.7 (693.2) & 882.3 (1213.0) & 302.5 (352.1) & 612.5 (813.0) \\
\bottomrule
\end{tabular}
\end{sc}
\end{small}
\end{table}
%%%%

We also compare the test inference times of our GPVAE model and the VLAE using the fully connected decoder networks. Note that VLAE is a semi-amortized approach that needs to perform the Laplace approximation at test time. Thus, another drawback is the computational overhead for inference,  which can be demanding as the number of linearization steps increases. The per-batch inference times (batch size 128) are shown in \autoref{tab:time_fullcon_decoder}. For the moderate or large linearization steps (e.g., 4 or 8), the inference takes significantly (about $2\times$ to over $3\times$) longer than that of our GPVAE (amortized method). 

%%%%
\begin{table}[t]
\centering
\caption{%(OMNIGLOT with 
(Fully connected networks as decoders) 
Per-batch inference time (unit in milliseconds) with batch size 128. %For the VLAE, we use the fully connected decoder with the same number of parameters as the convnet decoder. 
The figures without parentheses are the times using the fully connected networks, whereas figures in the parentheses are the times using the convolutional decoder networks. %Both architectures have roughly equal number of weight parameters.
}
\label{tab:time_fullcon_decoder}
\vskip 0.05in
\begin{small}
\begin{sc}
\centering
\begin{tabular}{l|cc|cc}
\toprule
 & \multicolumn{2}{c}{MNIST} & \multicolumn{2}{|c}{OMNIGLOT} \\
 \cline{2-5}
 & \ \ $\textrm{dim}({\bf z})=10$ \ \ & \ \ $\textrm{dim}({\bf z})=50$ \ \ & \ \ $\textrm{dim}({\bf z})=10$ \ \ & \ \ $\textrm{dim}({\bf z})=50$ \ \ \\
\midrule
VLAE ($1$) & 10.1 & 12.9 & 11.2 & 12.1 \\
VLAE ($2$) & 11.2 & 13.4 & 13.2 & 16.9 \\
VLAE ($4$) & 14.8 & 17.8 & 15.4 & 18.7 \\
VLAE ($8$) & 20.7 & 30.8 & 22.1 & 26.4 \\ \hline
GPVAE & 5.7 (5.9) & 9.8 (9.9) & 6.9 (6.9) & 10.2 (10.2) \\
\bottomrule
\end{tabular}
\end{sc}
\end{small}
\end{table}
%%%%

\end{document}